\documentclass[journal]{IEEEtran}
\usepackage{graphicx}
\usepackage{multicol}
\usepackage{subfigure}
\usepackage{amsmath}
\usepackage{amssymb}
\usepackage{amsfonts}
\usepackage{url}
\usepackage{booktabs}
\usepackage{makecell}
\usepackage{caption2}
\usepackage{bm}
\usepackage{pifont}
\usepackage{setspace}
\usepackage{color}
\usepackage{multirow}
\usepackage{algorithm}

\ifCLASSINFOpdf
\else
\fi

\begin{document}
\title{Adaptive Morphological Reconstruction for Seeded Image Segmentation}

\author{Tao~Lei,  %
        Xiaohong~Jia,\IEEEmembership{}
        Tongliang~Liu,\IEEEmembership{}
        Shigang~Liu,\IEEEmembership{~}
        Hongying~Meng,\IEEEmembership{~}
        and~Asoke~K.~Nandi\IEEEmembership{}

\thanks{

T.~Lei is with the  School of Electronical and Information Engineering, Shaanxi University of Science and Technology, Xi{'}an 710021, China, and also with the School of Computer Science, Northwestern Polytechnical University, Xi{'}an 710072, China. (E-mail: leitao@sust.edu.cn)

X.~Jia is with the  School of Electronical and Information Engineering, Shaanxi University of Science and Technology, Xi{'}an 710021, China. (E-mail: jiaxhsust@163.com)

T.~Liu is with the  School of Computer Science, The University of Sydney, Sydney, Australia. (E-mail: tliang.liu@gmail.com)

S.~Liu is with with the School of Computer Science, Shaanxi Normal University, Xi{'}an 710119, China. (E-mail: shgliu@gmail.com)

H.~Meng is with the Department of Electronic and Computer Engineering, Brunel University London, Uxbridge, Middlesex, UB8 3PH, U.K. (E-mail: hongying.meng@brunel.ac.uk)

A. K. Nandi is with the Department of Electronic and Computer Engineering, Brunel University London, Uxbridge, Middlesex, UB8 3PH, U.K., and also the Key Laboratory of Embedded Systems and Service Computing, College of Electronic and Information Engineering, Tongji University, Shanghai 200092, China. (E-mail: asoke.nandi@brunel.ac.uk)
}}

\markboth{}%
{Shell \MakeLowercase{\textit{et al.}}: Bare Demo of IEEEtran.cls for IEEE Journals}
\maketitle

\begin{abstract}
Morphological reconstruction (MR) is often employed by seeded image segmentation algorithms such as watershed transform and power watershed as it is able to filter seeds (regional minima) to reduce over-segmentation. However, MR might mistakenly filter meaningful seeds that are required for generating accurate segmentation and it is also sensitive to the scale because a single-scale structuring element is employed. In this paper, a novel adaptive morphological reconstruction (AMR) operation is proposed that has three advantages. Firstly, AMR can adaptively filter useless seeds while preserving meaningful ones. Secondly, AMR is insensitive to the scale of structuring elements because multiscale structuring elements are employed. Finally, AMR has two attractive properties: monotonic increasingness and convergence that help seeded segmentation algorithms to achieve a hierarchical segmentation. Experiments clearly demonstrate that AMR is useful for improving algorithms of seeded image segmentation and seed-based spectral segmentation. Compared to several state-of-the-art algorithms, the proposed algorithms provide better segmentation results requiring less computing time. Source code is available at {\color[rgb]{1,0,0.5}https://github.com/SUST-reynole/AMR}.
\end{abstract}

\IEEEpeerreviewmaketitle

\section{Introduction}\label{1}
\IEEEPARstart{M}{orphological} reconstruction (MR) [1] is a powerful operation in mathematical morphology. It has been widely used in image filtering [2], image segmentation [3], and feature extraction [4], etc. Among these applications, one of the most important applications is that MR is often used in seeded segmentation algorithms [5], [6] such as watershed transformation (WT) [7] and power watershed (PW) [8] to reduce over-segmentation caused by image noise and details. However, there are two drawbacks [9], [10] when MR is used in seeded segmentation algorithms.

\begin{itemize}
\item It is difficult to reduce over-segmentation while obtaining a high segmentation accuracy for seeded segmentation algorithms (we use MR-WT to denote MR-based watershed transform and use MR-PW to denote MR-based power watershed). Although MR is able to filter noise in gradient images, some important contour details are smoothed out as well.

\item MR is sensitive to the scale of structuring elements. In practical applications, if the scale is too small, the reconstructed gradient image suffers from a serious over-segmentation. Oppositely, if the scale is too large, the reconstructed gradient image suffers from an under-segmentation.
\end{itemize}

Generally, MR is used in watershed transform to improve the segmentation effect by employing a structuring element to filter regional minima [11]. However, it is very difficult to filter useless regional minima while preserving meaningful ones by simply considering one single-scale structuring element. Although $H$-min imposition [12] is a simple and efficient method for over-segmentation reduction, it relies on a threshold choice and is likely to miss some important boundaries. Region merging [13], [14] is also a popular method for this, but it requires iterating and renewing edge weight leading to a high computing burden. In addition, some researchers employ reasonable contour detection methods, e.g., globalized probability of boundary (gPb) [15] that combines the multiscale information from brightness, color and texture, to achieve better image segmentation. However, the gPb is computationally expensive because it combines too many feature cues for contour detection. To speed up the algorithm of contour detection, Dollar and Zitnick [16] took the advantage of the structure present in regional image patches and random decision forests, and proposed a fast structured edge (SE) detection approach using structured forests. This algorithm obtains real-time performance and state-of-the-art edge detection but requires a huge amount of memory for training data. To reduce memory requirement, Hallman and Fowlkes [17] proposed a simple and efficient model to learn contour detection, namely oriented edge forests (OEF). Although these improved contour detectors are superior to traditional detectors, e.g. Sobel or Canny, and they are helpful for improving subsequent image segmentation, they still generate a large number of seeds leading to serious over-segmentations.

In practice, contour detection methods are usually combined with other approaches to improve image segmentation effect. For example, Fu \emph{et al.}  [18] proposed a robust image segmentation approach using contour-guided color palettes  by integrating contour and color cues, where SE, mean-shift algorithm [19], region merging, and spectral clustering [20] are combined to achieve better segmentation results. However, the approach is complex because it combines several different algorithms that requires many parameters.

In this paper, we propose an adaptive morphological reconstruction (AMR) operation that is able to generate a better seed image than MR to improve seeded segmentation algorithms. Firstly, AMR employs multiscale structuring elements to reconstruct a gradient image. Secondly, a pointwise maximum operation on these reconstructed gradient images is performed to obtain the final adaptive reconstruction result. Because AMR employs small structuring elements to reconstruct pixels of large gradient magnitudes while employing large structuring elements to reconstruct pixels of small gradient magnitudes in a gradient image, AMR is able to obtain better seed images to improve the seeded segmentation algorithms. Our main contributions are summarized as follows.
\begin{itemize}
\item Multiscale structuring elements are employed by AMR, and different scaled structuring elements are adaptively adopted by pixels of different gradient magnitudes without computing the local features of a gradient image.

\item AMR has a convergence property and a monotonic increasing property, the two properties help seeded segmentation algorithms to achieve a hierarchical segmentation.

\item AMR has a low computational complexity, and it can help seed-based spectral segmentation to achieve better image segmentation results than the-state-of-art algorithms.
\end{itemize}

The rest of the paper is organized as follows. In the next section, the research background related with AMR is introduced and analyzed. On this basis, AMR is proposed, and its two properties, monotonic increasingness and convergence are carefully analyzed in Section \uppercase\expandafter{\romannumeral 3}. To demonstrate the superiority of AMR, AMR is used for seeded image segmentation and seed-based spectral segmentation. Experiments are presented in Section  \uppercase\expandafter{\romannumeral 4}, followed by the conclusion in Section \uppercase\expandafter{\romannumeral 5}.
\begin{figure}[t]\label{figure 1}  
\renewcommand{\captionlabeldelim}{.}
  \centering
  \subfigure[]{
   \begin{minipage}[b]{0.5\textwidth}
     \centering
      \includegraphics[width=0.8\textwidth]{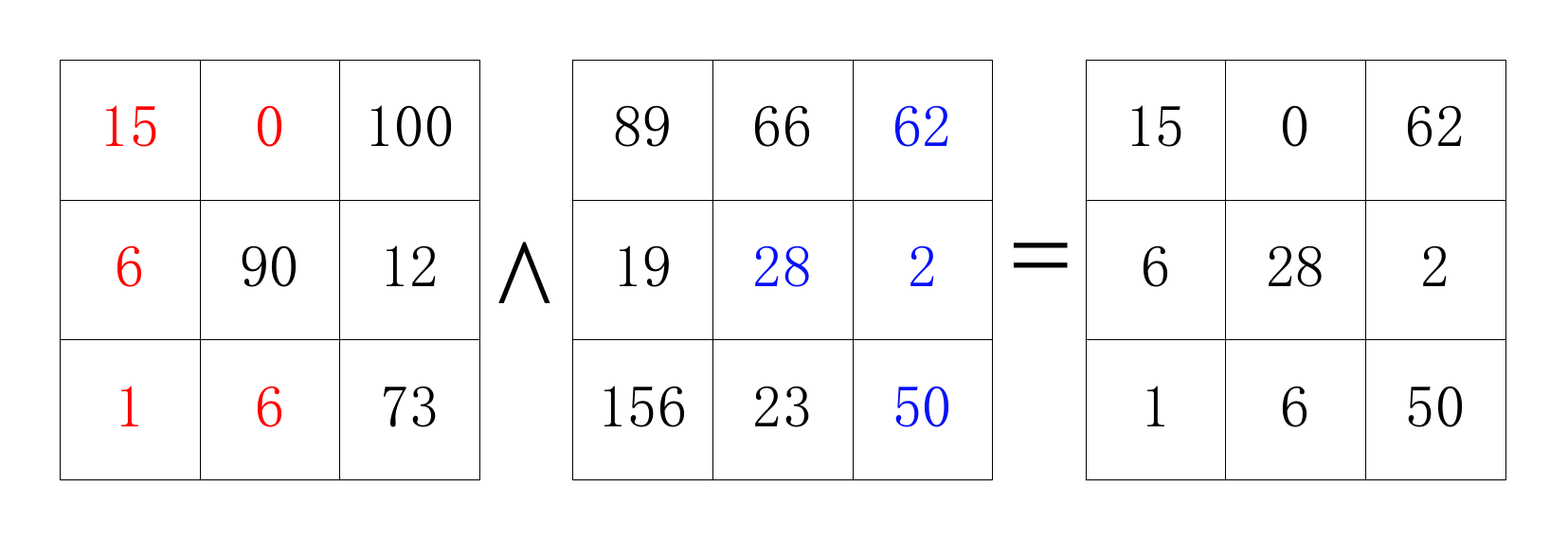}
       \vspace{-0.2cm}
       \end{minipage}
  }
 \vspace{-0.3cm}

  \subfigure[]{
      \begin{minipage}[b]{0.5\textwidth}
        \centering
      \includegraphics[width=0.8\textwidth]{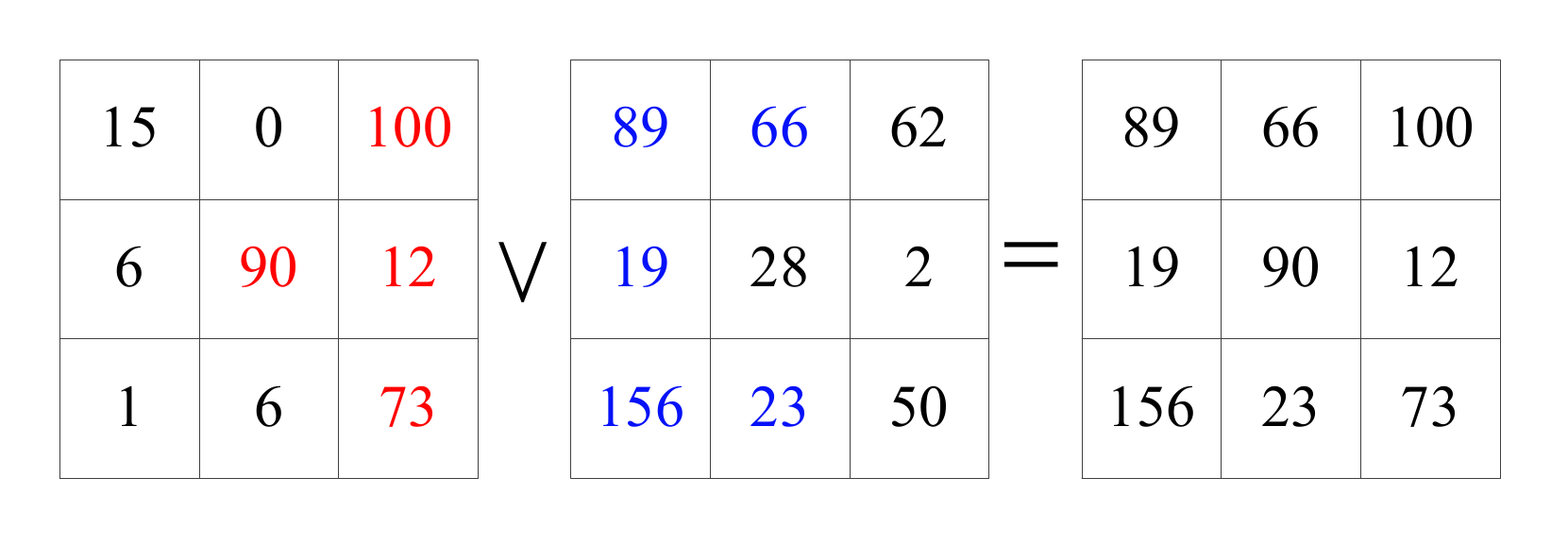}
        \vspace{-0.2cm}
      \end{minipage}
  }
  \vspace{-0.01cm}
  \captionsetup{font={small}}
  \caption{An example for pointwise extremum operation. (a) Pointwise minimum. (b) Pointwise maximum.}
  \end{figure}

 \begin{figure}[t]\label{fig.2}
\renewcommand{\captionlabeldelim}{.}\small
	    \begin{center}
	        \includegraphics[width=0.65\linewidth]{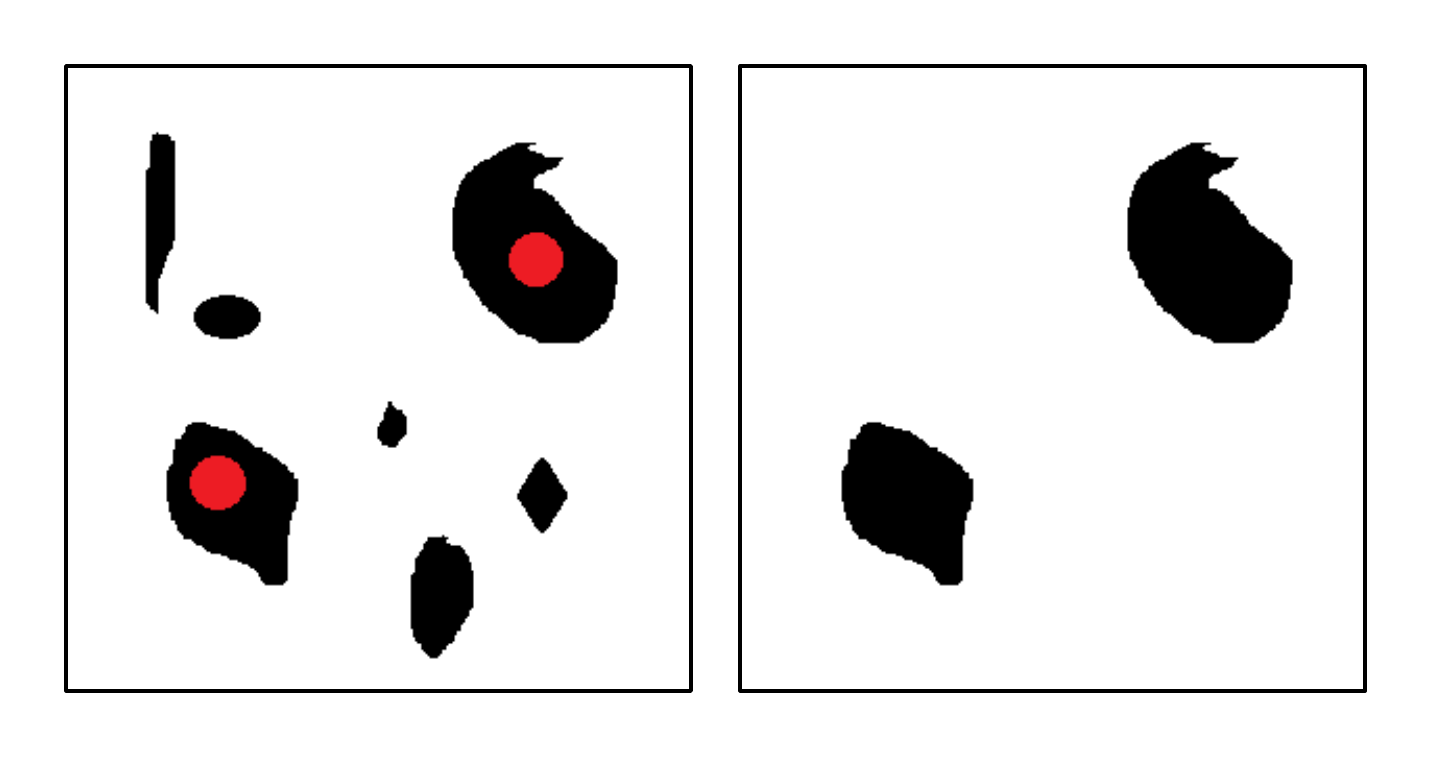}
             \centerline{(a)~~~~~~~~~~~~~~~~~~~~~(b)}
    	\end{center}
     \vspace{-0.2cm}
	\captionsetup{font={small}}\caption{Binary MR from markers. (a) A mask image. (b) Reconstructed result.}
\end{figure}

\begin{figure}[t]\label{fig.3}
\renewcommand{\captionlabeldelim}{.}\small
	    \begin{center}
	        \includegraphics[width=0.90\linewidth]{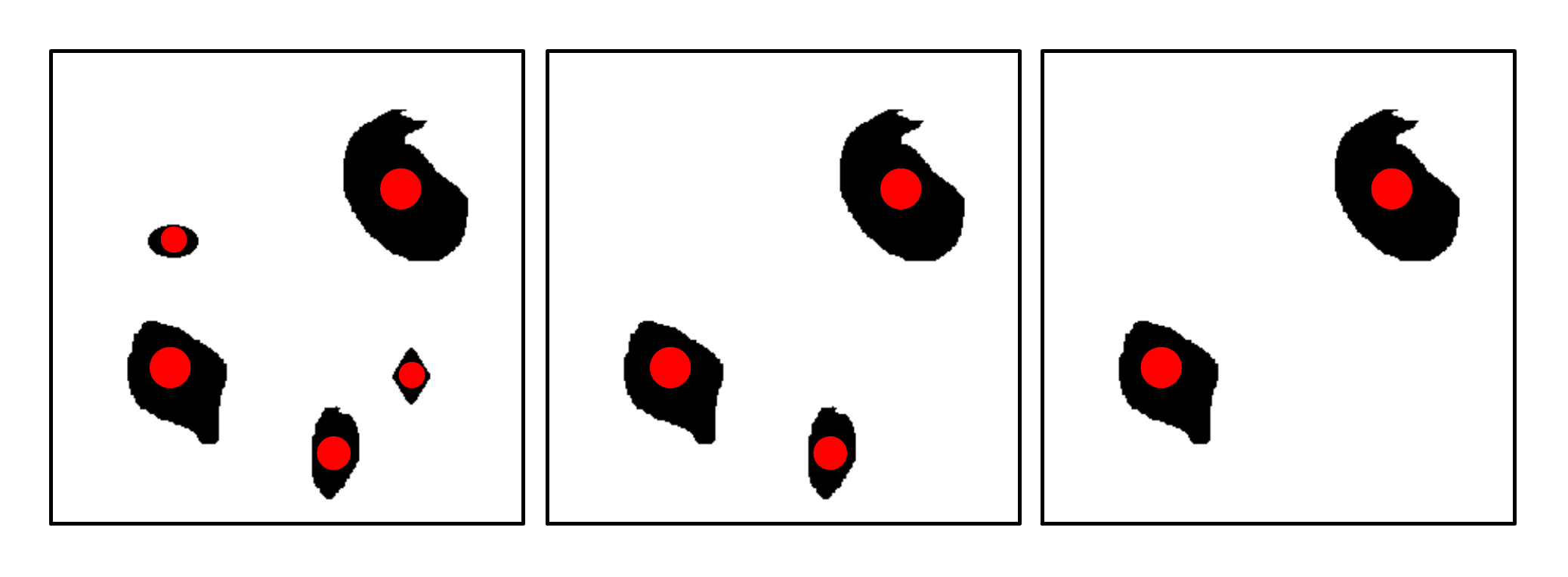}
            \centerline {(a) Mask 1.~~~~~~~~(b) Mask 2.~~~~~~~~(c) Mask 3.}
    	\end{center}
     \vspace{-0.2cm}
	\captionsetup{font={small}}\caption{Binary MR from different markers.}
\end{figure}

\section{Background}  %
\subsection{Morphological Reconstruction}

MR is an image transformation that requires two input images, a marker image and a mask image. Let two grayscale images $f$ and $g$ denote the marker image that is the starting point for the transformation and the mask image that constrains the transformation, respectively [21]. If $f\leq g$, which means $f$ is pointwise less than or equal to $g$, the morphological dilation reconstruction $(R^{\delta})$ of $g$ from $f$ is denoted by
\begin{equation}\label{eq.1}
  R_{g}^{\delta}(f)=\delta_{g}^{(n)}(f),
\end{equation}
where $\delta_{g}^{(1)}(f)=\delta(f)\wedge g$, $\delta_{g}^{(k)}(f)=\delta(\delta_{g}^{(k-1)}(f))\wedge g$ for $2\leq k\leq n$, $k, n\in N^{+}$ satisfies $\delta_{g}^{(n)}(f)=\delta_{g}^{(n-1)}(f)$. The symbol $\delta$  represents the elementary morphological dilation operation, and $\wedge$ stands for the pointwise minimum at each pixel of two images as shown in Fig. 1(a).

Similarly, if $f\geq g$, the morphological erosion reconstruction $(R^{\varepsilon})$ of $g$ from $f$, which is the dual operation of $R^{\delta}$, is defined as
\begin{equation}\label{eq.2}
R_{g}^{\varepsilon}(f)=\varepsilon_{g}^{(n)}(f),
\end{equation}
where $\varepsilon_{g}^{(1)}(f)=\varepsilon(f)\vee g$, $\varepsilon_{g}^{(k)}(f)=\varepsilon(\varepsilon_{g}^{(k-1)}(f))\vee g$ for $2\leq k\leq n$, $k, n\in  N^{+}$ satisfies $\varepsilon_{g}^{(n)}(f)=\varepsilon_{g}^{(n-1)}(f)$. The symbol $\varepsilon$ represents the elementary morphological erosion operation, and $\vee$ stands for the pointwise maximum at per pixel of two images as shown in Fig. 1(b).

To further illustrate the principle of MR for image transformation, we present an example for the binary MR as shown in Fig. 2, where the red regions denote seeds, i.e., the marker image $f$.

According to Fig. 2 and (1)-(2), a suitable marker image is important for MR. We have known that $f\leq g$ for $R^{\delta}$ while $f\geq g$ for $R^{\varepsilon}$. Thus, there are lots of choices for $f$. Different marker images corresponds to different reconstruction results as shown in Fig. 3. To obtain an efficient $f$ in practice, the marker image is usually obtained by performing a transformation on the corresponding mask image [22]-[24]. For example, the erosion or dilation result of a mask image is often considered as a marker image [25], i.e., $f=\varepsilon_{b_{i}}(g)$ or $f=\delta_{b_{i}}(g)$, where $b_{i}$ is a disk shaped structuring element, the radius of $b_{i}$ is $i$, $i\in N^{+}$. Therefore, MR is sensitive to the parameter $i$ because the marker image is decided by the scale of the structuring element.

As compositional morphological opening and closing operations show better performance than elementary morphological erosion and dilation operations for image filtering, feature extraction, etc., we present the definition of compositional morphological opening and closing reconstructions ($R^\gamma$ and $R^\phi$) of $g$ from $f$ as follows
\begin{equation}\label{eq.3}
\left\{
\begin{aligned}
R_{g}^{\gamma}(f)=R_{g}^{\delta}(R_{g}^{\varepsilon}(f))\\
R_{g}^{\phi}(f)=R_{g}^{\varepsilon}(R_{g}^{\delta}(f))
\end{aligned}.
\right.
\end{equation}

\begin{figure}[t]\label{fig.4}
\renewcommand{\captionlabeldelim}{.}\small
	    \begin{center}
	        \includegraphics[width=1.0\linewidth]{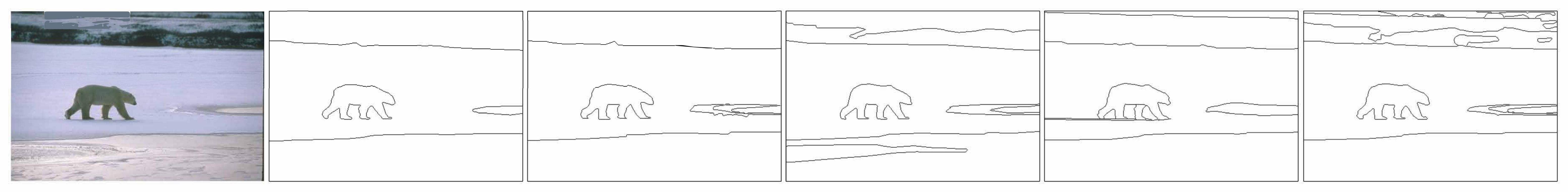}
            \centerline{~~(a)~~~~~~~~~~(b)~~~~~~~~~~(c)~~~~~~~~~~(d)~~~~~~~~~~(e)~~~~~~~~~~(f)~~}

    	\end{center}
     \vspace{-0.1cm}  %
	\captionsetup{font={small}}\caption{The original image and ground truths (GT) from BSDS500. (a) ``100007''. (b) GT 1. (c) GT 2.  (d) GT 3.  (e) GT 4. (f) GT 5. BSDS (http://www.eecs.berkeley.edu/Research/Projects/CS/vision/bsds/) is a very popular image dataset and it is often used for the evaluation of image segmentation algorithms. For each image in BSDS, there are 4 to 9 ground truths segmentations that are delineated by different human subjects.}
\end{figure}

\subsection{Multiscale and Adaptive Mathematical Morphology}\label{2.1} 
For image filtering and enhancement using morphological operators, a large-scale structuring element can suppress noise but may also blur the image details, whereas a small-scale structuring element can preserve image details but may fail to suppress noise. Some researchers proposed multiscale and adaptive morphological operators to improve the performance of traditional morphological operators. However, most multiscale morphological operators [26], [27] such as morphological gradient operators and morphological filtering operators, average all scales of morphological operation results as final output
\begin{equation}\label{eq.4}
  y=\frac{1}{\lambda}\sum_{j=1}^{\lambda}g_j,
\end{equation}
where $y$ is the final output result, $j$ is the radius of the structuring element, $1\leq j\leq \lambda$, $j$, $\lambda\in N^+$. Although the average result is superior to the result based on single-scale morphological operators, it causes contour offset and mistakes. Some researchers improved multiscale morphological operators by introducing a weighted coefficient to (4), and they defined adaptive multiscale morphological operators as follows [28]
\begin{equation}\label{eq.5}
  y=\frac{1}{\lambda}\sum_{j=1}^{\lambda}\omega_{j}g_{j},
\end{equation}
where $w_j$ is the weighted coefficient on the $j$th scale result. However, because the computing of weighted coefficients is complex, the adaptive multiscale morphological operators have a low computational efficiency. Moreover, the weighted average result is similar to average result because it is difficult to obtain the optimal weighted coefficient, even though the former is slightly better than the latter.

Although lots of adaptive multiscale morphological operators [29]-[32] have been proposed, it can be seen from (4)-(5) that both the multiscale and adaptive morphological operators employ a linear combination of different-scale results to improve single-scale morphological gradient or filtering operators. Because the linear combination is unsuitable for multiscale morphological reconstruction operation, in this paper, we try to employ a non-linear combination (i.e., the pointwise maximum operation denoted by $\vee$) to design adaptive morphological reconstruction operators. These operators are different from conventional multiscale and adaptive morphological operators employing linear combination in (4)-(5). We use non-linear operation $\vee$ instead of linear combination since the former is more suitable than the later for the removal of useless seeds in seeded image segmentation.

\begin{figure}[t]\label{fig.5}
\renewcommand{\captionlabeldelim}{.}\small
	    \begin{center}
	        \includegraphics[width=1.0\linewidth]{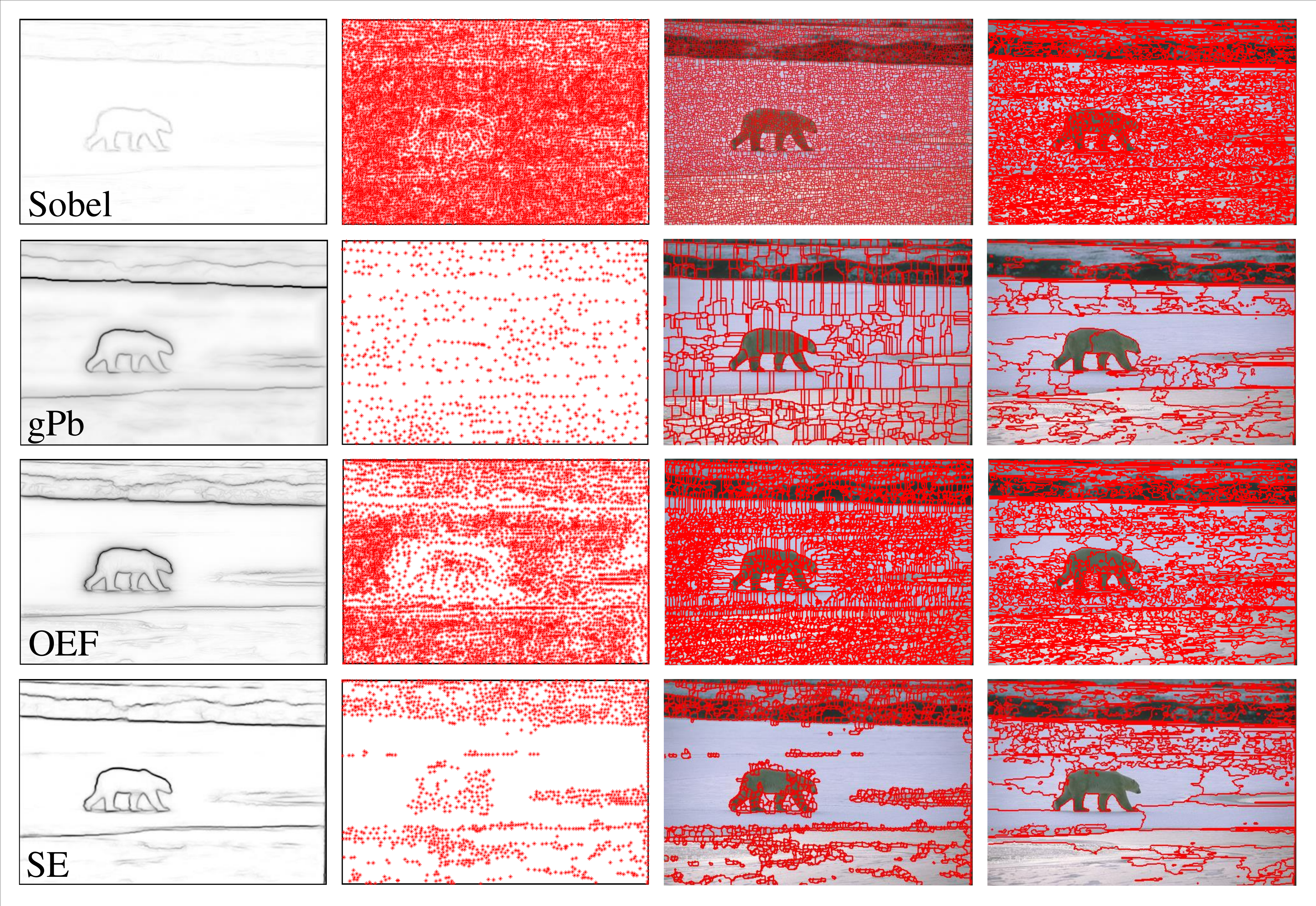}
             {(a)~~~~~~~~~~~~~~~~~(b)~~~~~~~~~~~~~~~~~(c)~~~~~~~~~~~~~~~~(d)}
    	\end{center}
     \vspace{-0.1cm}  %
	\captionsetup{font={small}}\caption{Over-segmentation reduction by improving the gradient image of ``100007''. (a) Different gradient images. (b) Seed images (regional minima). (c) WT. (d) PW ($p=2$) [8].}
\end{figure}

\begin{figure*}[t]\label{fig.6}
\renewcommand{\captionlabeldelim}{.}\small
	    \begin{center}
	        \includegraphics[width=1.0\linewidth]{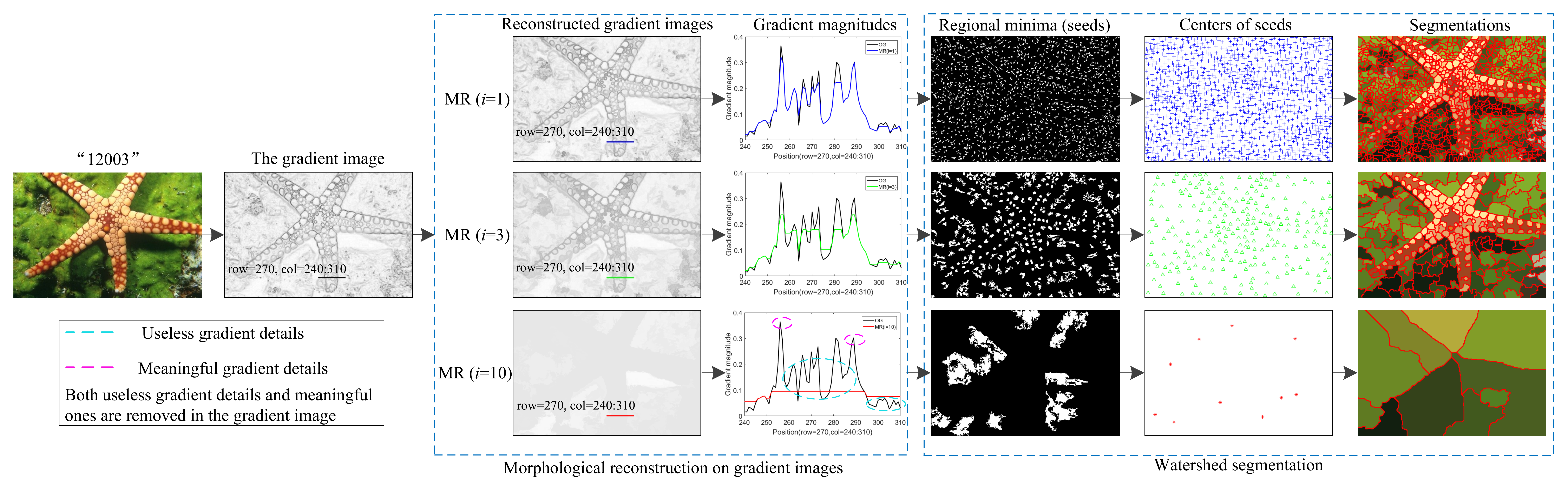}
    	\end{center}
     \vspace{-0.1cm}  %
	\captionsetup{font={small}}\caption{A seeded segmentation framework based on MR-WT. ($R_g^\phi(f)$ is employed to reconstruct a gradient image, and original gradient (OG) denotes a row of the original gradient image).}
\end{figure*}

\begin{figure}[t]\label{fig.7}
\renewcommand{\captionlabeldelim}{.}\small
	    \begin{center}
	        \includegraphics[width=0.9\linewidth]{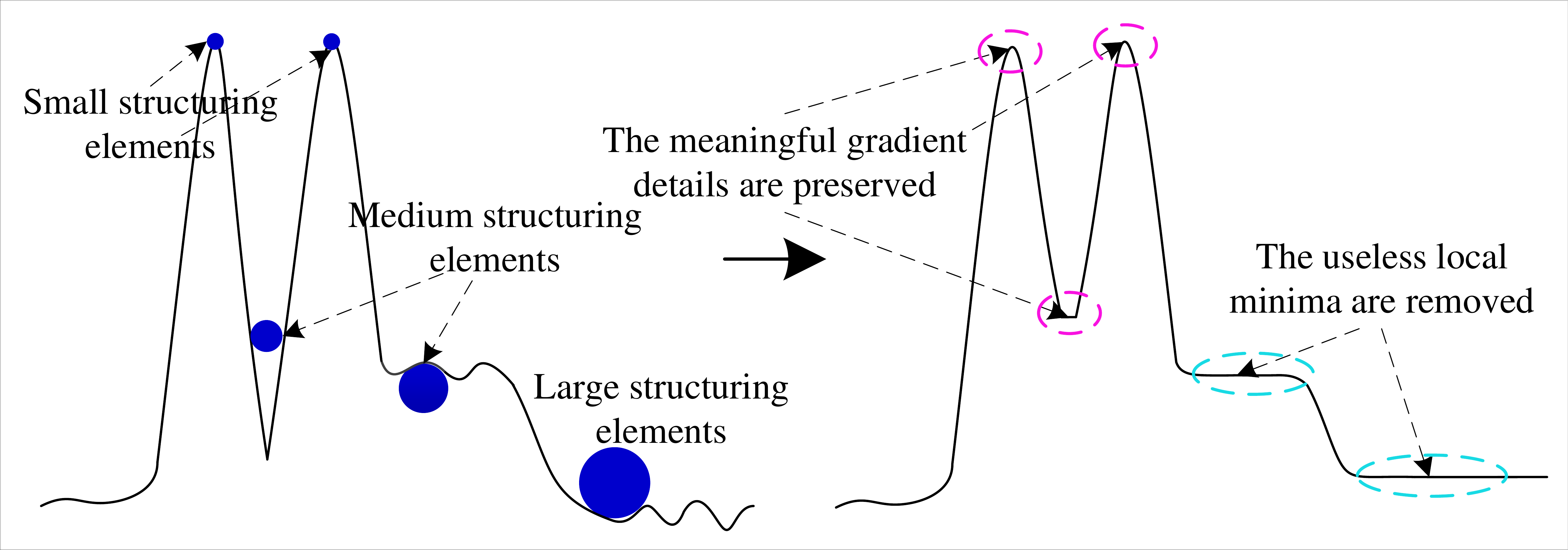}
              {(a)~~~~~~~~~~~~~~~~~~~~~~~~~~~~~~~(b)}
    	\end{center}
     \vspace{-0.1cm}
	\captionsetup{font={small}}\caption{The motivation of AMR. (a) Gradient. (b) Reconstructed gradient.}
\end{figure}

\begin{figure*}[t]\label{fig.8}
\renewcommand{\captionlabeldelim}{.}\small
	    \begin{center}
	        \includegraphics[width=1.0\linewidth]{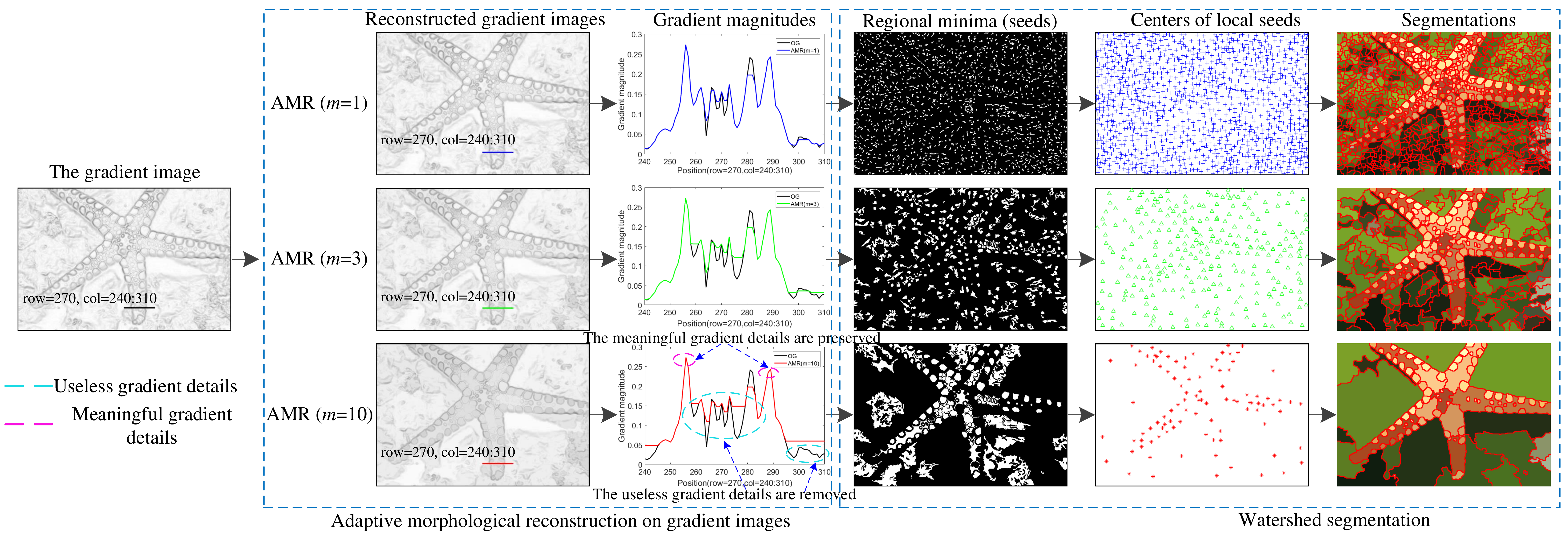}
    	\end{center}
     \vspace{-0.1cm}
	\captionsetup{font={small}}\caption{Seeded segmentation framework based on AMR-WT.}
\end{figure*}

\begin{figure}[t]\label{fig.9}
\renewcommand{\captionlabeldelim}{.}\small
	    \begin{center}
	        \includegraphics[width=1.0\linewidth]{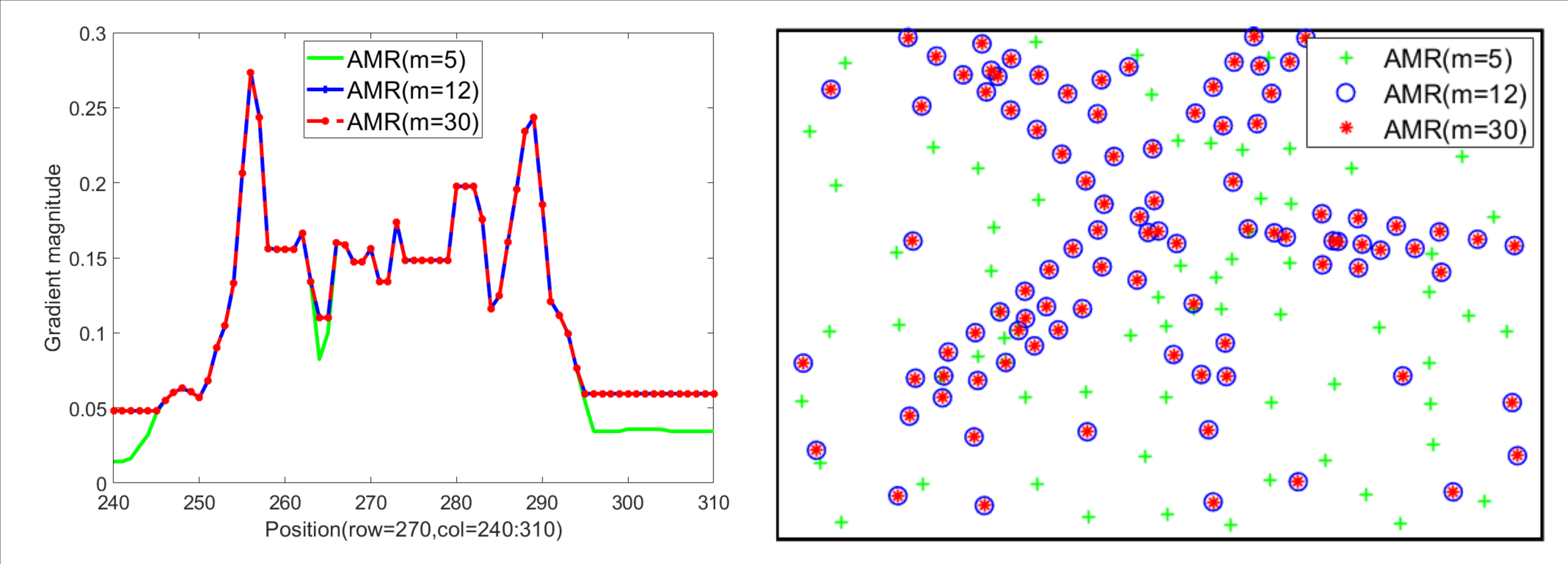}
            {(a)~~~~~~~~~~~~~~~~~~~~~~~~~~~~~~~~(b)}
    	\end{center}
     \vspace{-0.1cm}
	\captionsetup{font={small}}\caption{Comparison of gradient reconstruction and seed filtering with variant value of $m$. Because AMR has an important property of convergence, the seed image is unchanged when the value of $m$ is large enough. The seed image is unchanged when m$\geq$12 for the image ``12003''. (a) The variation of gradient magnitudes. (b) The variation of seed images.}
\end{figure}

\subsection{Seeded Segmentation}\label{2.3} %
Seeded segmentation algorithms, such as graph cuts [33], random walker [34], watersheds [7], and power watershed [8] have been widely used in complex image segmentation tasks due to their good performance [35]. It is not required to give seed images for both graph cuts and random walker because they usually consider each pixel as a seed. However, a seed image is necessary for WT and PW by computing the regional minima of a gradient image.

Since both WT and PW obtain seeds from a gradient image that often includes a huge number of seeds generated by noise and unimportant texture details, they usually suffer from over-segmentation. A larger number of approaches for addressing over-segmentation was proposed, and these approaches can be categorized into two groups.

\begin{itemize}
\item Feature extraction or feature learning is used to obtain a better gradient image that enhances important contours while smoothing noise and texture details [15]-[17].
\item MR is used for gradient reconstruction to reduce the number of regional minima [36]-[38].
\end{itemize}

For the first group of approaches, gPb, OEF, and SE are popular for reducing over-segmentation as shown in Figs. 4-5. In Fig. 5, although gPb, OEF, and SE provide better gradient images that can reduce over-segmentation for WT and PW, the segmentation results are still poor compared to ground truths shown in Fig. 4.

The second group of approaches depends on MR and WT, it is denoted by MR-WT. Najman and Schmitt [36] employed MR to remove regional minima to reduce over-segmentation. Furthermore, a dynamic threshold is used to change the gradient magnitude that is smaller than the threshold, and then a hierarchical segmentation result is obtained. Wang [26] proposed a multiscale morphological gradient algorithm (MMG) for image segmentation using watersheds. The proposed MMG employs multiple structuring elements to obtain a better gradient image, and uses MR to remove regional minima to improve watershed segmentation.

Fig. 6 illustrates the principle of a seeded segmentation framework based on MR-WT. We can see that the number of the regional minima in the gradient image decreases rapidly with the increase of $i$, but the boundary is also destroyed simultaneously. It is clear that the larger structuring element corresponds to fewer seeds. One major reason is that MR employs a single-scale structuring element, which equally treats all pixels of different gradient magnitudes in the gradient image. For example, in dilation reconstruction, the marker image $f=\varepsilon_{b_i}(g)$ converges to the minimum grayscale value of pixels in the mask image as the value of $i$ increases. Obviously, both large and small structuring elements lead to poor reconstruction results while a moderate-sized structuring element achieves a rough balance via sacrificing contour precision. Therefore, it is difficult to obtain a good seed image by employing a single-scale structuring element. Although many researchers employ multiscale structuring elements to generate a better gradient image, there are few studies on multiscale MR for gradient images. Moreover, the fusion of different-scale results is also a problem.
\vspace{-0.2cm}
\subsection{Spectral Segmentation}
It is well-known that spectral clustering [20] is greatly successful due to the fact that it does not make strong assumptions on data distribution, and it is implemented efficiently even for large datasets, as long as we make sure that the affinity matrix is sparse. However, since the size of the affinity matrix is $(M\times N)^2$ for an image of size $M\times N$, and it is not sparse because of Gaussian similarity measure, spectral clustering is often inefficient for image segmentation due to eigenvalue decomposition of the huge affinity matrix. To address the issue, a great number of algorithms have been proposed to construct a smaller affinity matrix and thus to improve the computational efficiency of spectral clustering [39]-[42]. Most of these algorithms employ pre-segmentation (superpixel) methods such as the simple linear iterative clustering (SLIC) [43], mean-shift [19], linear spectral clustering (LSC) [44], and superpixel hierarchy [45], to reduce the number of pixels of the original image and, in turn, reduces the size of the affinity matrix. As an example, Zhang \emph{et al.} [46] proposed a fast image segmentation approach that is a re-examination of spectral clustering on image segmentation. The approach provides better image segmentation results yet requires a long running time.

The popular superpixel approaches have some drawbacks for spectral segmentation. Firstly, mean-shift algorithm involves three parameters and it is sensitive to these parameters. Secondly, SLIC only generates superpixels that include regular regions, and these regions have a similar shape and size. Finally, LSC is superior to SLIC because LSC successfully connects a local feature with a global optimization objective function, so that LSC can generate more reasonable segmentation results. However, similar to SLIC, LSC also provides superpixels that include regular regions with a similar shape and size.

As seed-based spectral segmentation algorithms are sensitive to pre-segmentation results, an excellent pre-segmentation algorithm can improve segmentation results generated by seed-based spectral segmentation algorithms.

\begin{figure}[t]\label{fig.10}
\renewcommand{\captionlabeldelim}{.}\small
	    \begin{center}
	        \includegraphics[width=1.0\linewidth]{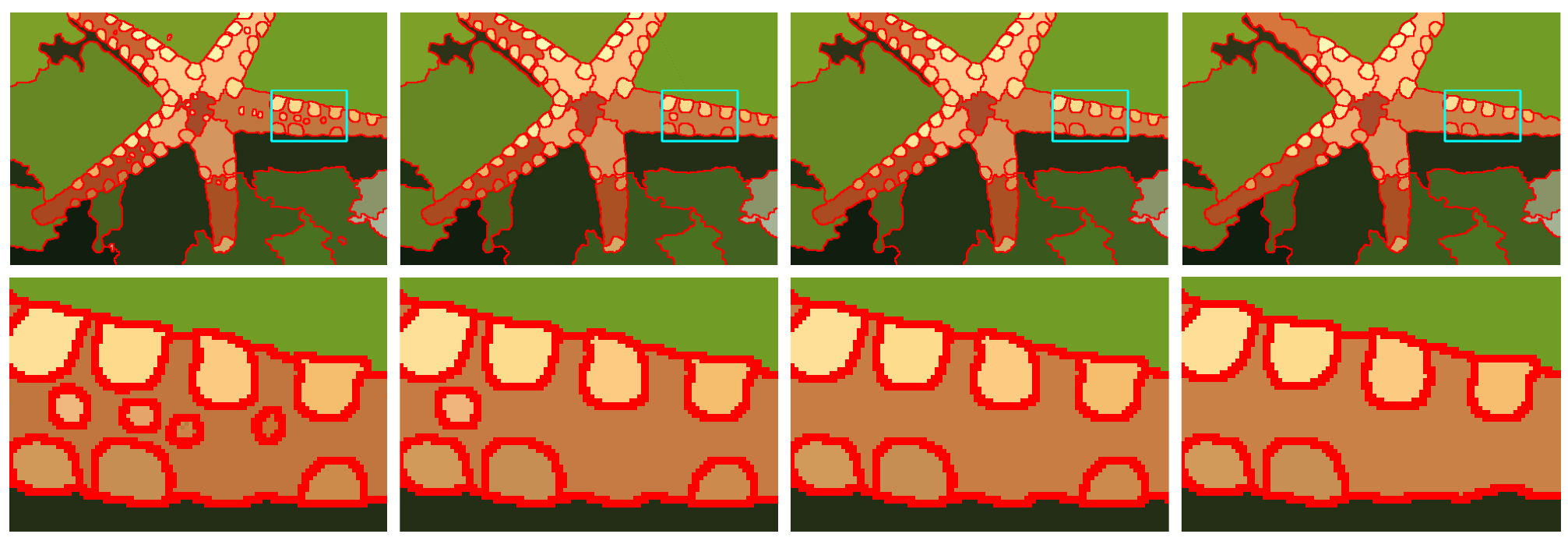}
 {(a)~~~~~~~~~~~~~~~~(b)~~~~~~~~~~~~~~~~(c)~~~~~~~~~~~~~~~~(d)}
    	\end{center}
     \vspace{-0.2cm}
	\captionsetup{font={small}}\caption{Segmentation results using AMR-WT by changing the value of $s$. (a) $s=1$, $m=10$. (b) $s=2$, $m=10$. (c) $s=3$, $m=10$. (d) $s=5$, $m=10$.}
\end{figure}
\vspace{-0.1cm}

\section{Adaptive morphological reconstruction}

\subsection{The Proposed AMR}
To overcome the drawback of MR on regional minima filtering, we propose an AMR that is able to filter useless regional minima and maintains meaningful ones generated by salient objects. Fig. 7 shows the motivation of AMR in which multiscale structuring elements are employed to reconstruct a gradient image, i.e., small structuring elements are adopted by pixels of large gradient magnitude while large structuring elements are adopted by pixels of small gradient magnitude.

\textbf{Definition 1.} Let $b_{s}\subseteq\cdots b_{i}\subseteq b_{i+1}\cdots\subseteq b_{m}$ be a series of nested structuring elements, where $i$ is the scale parameter of a structuring element, $1\leq s \leq i \leq m$, $s,i,m \in N^+$. For a gradient image $g$ such that $f=\varepsilon_{b_{i}}(g)$ and $f\leq g$, the adaptive morphological reconstruction denoted by $\psi$ of $g$ from $f$ is defined as
\begin{equation}\label{eq.6}
\psi(g,s,m)=\vee_{s\leq i\leq m}\left\{R_{g}^{\phi}(f)_{b_{i}}\right\}.
\end{equation}

Note that the pointwise maximum operation is only suitable for $R^\phi$, but not suitable for $R^\gamma$. Because $\lim\limits_{m\rightarrow\infty}R_g^{\gamma}(f)_{b_m}=max(g)$ (the proof is presented in Appendix A) and $\lim\limits_{m\rightarrow\infty}\vee_{s\leq i \leq m}\left\{R_g^\gamma(f)_{b_i}\right\}=max(g)$, $\psi(g,s,m)$ is unable to obtain a significantly convergent gradient image if $\psi(g,s,m)=\vee_{s\leq i \leq m}\left\{R_g^{\gamma}(f)_{b_i}\right\}$.

We apply AMR to the gradient image shown in Fig. 6. The reconstruction and segmentation results are shown in Fig. 8, where the adopted structuring elements are disk and $s=1$. More detailed comparisons are shown in Fig. 9. By comparing Fig. 8 with Fig. 6, it is obvious that AMR obtains better seed images than MR due to the fact that the non-linear operation $\vee$ is able to remove efficiently useless seeds.

To further show the influence of $s$ on AMR, Fig. 10 shows segmentation results provided by AMR through changing the value of $s$. We can see that there are some small segmented areas when the value of $s$ is small. These small areas are merged by increasing the value of $s$. However, although a large $s$ leads to the merge of small areas, the precision of object contours will be decreased as shown in Fig. 6. Therefore, we usually set $1\leq s \leq 3$ for a moderate-sized image.

\subsection{The Monotonic Increasingness Property of AMR}
AMR is an algorithm that aims at finding meaningful regional minima by merging or filtering useless regional minima. AMR includes two parameters $s$ and $m$. When we increase the value of $m$, gradient images reconstructed by AMR keep the increasing order as shown in Theorem 1.

\textbf{Theorem 1.} Let $\psi$ be an adaptive morphological reconstruction operator, $\psi$ is increasing with respect to the scale of structuring elements, i.e., for a gradient image $g$ such that $f=\varepsilon_{b_i}(g)$ and $f\leq g$, $1\leq p$, $q \leq m$, $p,q,m\in N^+$, we have
\begin{equation}\label{eq.7}
  p\leq q\Rightarrow\psi(g,s,p)\leq\psi(g,s,q).
\end{equation}

The proof of Theorem 1 is presented in Appendix B. Theorem 1 shows that the gradient image processed by AMR is monotonous increasing with the increase of $m$. Fig. 9 demonstrates Theorem 1. We can see that if $m$ is enlarged, the more unimportant seeds are removed, and important seeds are preserved. Actually, the result is equivalent to region merging. However, the method is simpler than region merging. According to the result, it can be seen that AMR can help seeded segmentation algorithms to achieve a hierarchical segmentation [47], [48]. Hierarchical segmentation is a multilevel segmentation scheme, and it usually outputs a coarse-to-fine hierarchy of segments ordered by the level of details. Multiscale combinatorial grouping (MCG) proposed by Pont-Tuset \emph{et al.} [49] is an excellent hierarchical segmentation approach that employs a fast normalized cut algorithm and an efficient algorithm for combinatorial merging of hierarchical regions. Based on the hierarchical segmentation results provided by MCG, some improved approaches are also proposed [50], [51]. These improved approaches achieve better segmentation effect but have lower computational efficiency than MCG.

Before analyzing the relationship between AMR-WT and hierarchical segmentation, we first review some basic concepts of hierarchical segmentation. Let $\Omega$ be a finite set. A hierarchy $H$ on $\Omega$ is a set of parts of $\Omega$ such that
\begin{itemize}
  \item $\Omega\in H$.
  \item For every $\omega\in \left\{\Omega\right\}$, $\left\{\omega\right\}\in H$.
  \item For each pair $(h, h')\in H^2$, $h\bigcap h'\neq\varnothing\Rightarrow h\subseteq h' $ or $ h' \subseteq h$.
\end{itemize}

Note that $H$ is a chain of nested partitions. Let $H_0$ be the initial partition of $\Omega$, which corresponds to the finest partition of $\Omega$, and $H_n$ be the coarsest partition of $\Omega$, which segments the images as one single region. A partition $H_z$, $0\leq z \leq n$, on $\Omega$ has the property that
\begin{equation} \label{eq.8}
H_z = H_0, if~z\leq0,
\end{equation}
\begin{equation}\label{eq.9}
\exists n\in N^+, H_z=\left\{\Omega\right\}, \forall z\geq n,
\end{equation}
\begin{equation}\label{eq.10}
p\leq q\Rightarrow H_p \subseteq H_q, 1\leq p, q < n,
\end{equation}
where $H_p \subseteq H_q$ denotes the partition. $H_p$ is finer than the partition $H_q$. Derived from Theorem 1 and Fig. 9, we obtain
\begin{equation}\label{eq.11}
\psi(g,s,p)\leq \psi(g,s,q) \Rightarrow S(\psi(g,s,p))\subseteq S(\psi(g,s,q)),
\end{equation}
where $S$ denotes seeded segmentation algorithms such as WT or PW. Suppose that $H_0=S(g)$, $H_m=S(\psi(g,s,m))$, and $s=1$ then
\begin{equation} \label{eq.12}
H_0\subseteq H_1\subseteq\cdots\subseteq H_m.
\end{equation}

According to (12), the principle of the hierarchical segmentation based on AMR is shown in Fig. 11, in which the data points represent regions obtained by the hierarchical segmentation at different levels.

\begin{figure}[t]\label{fig.11}
\renewcommand{\captionlabeldelim}{.}\small
	    \begin{center}
	        \includegraphics[width=0.65\linewidth]{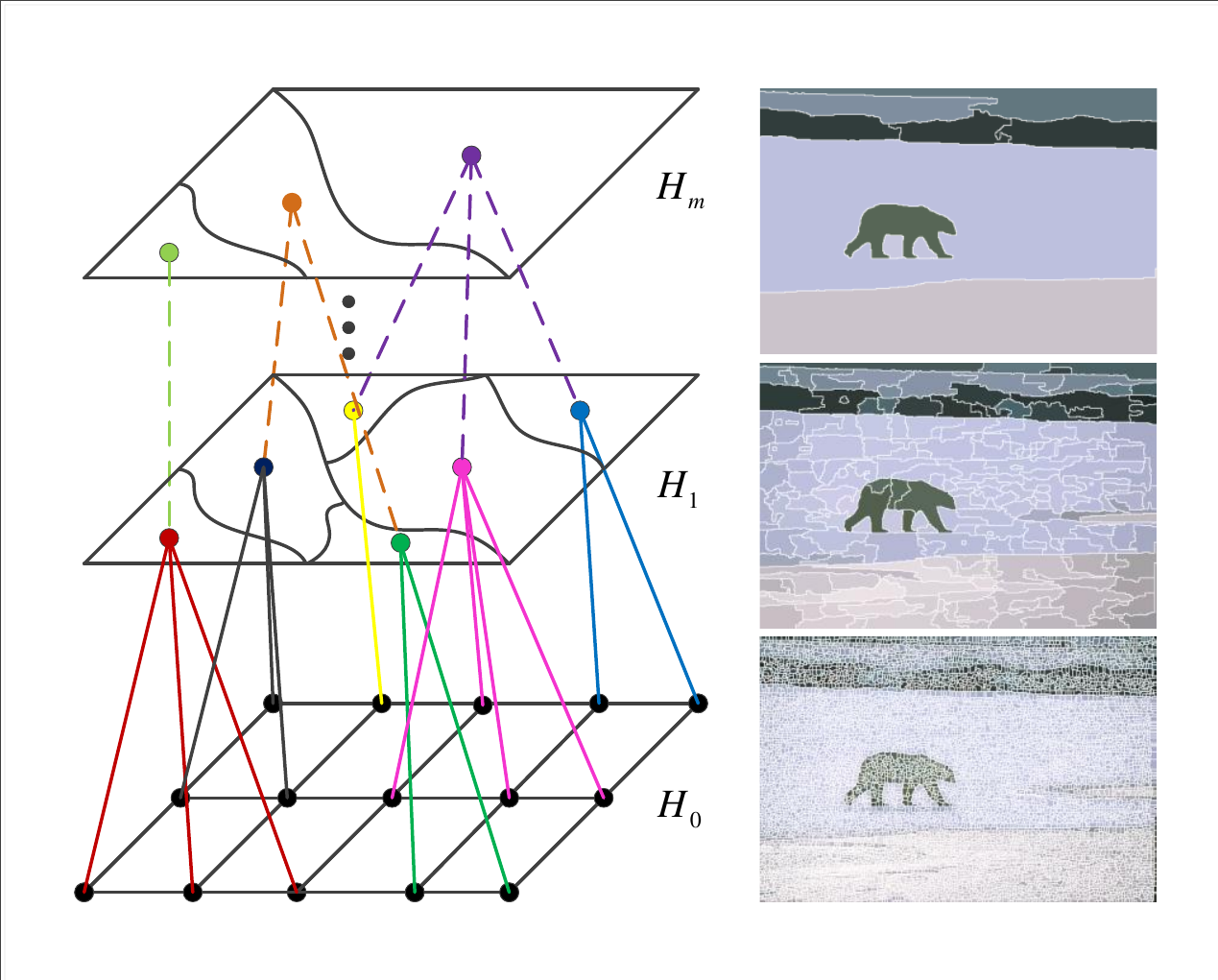}
    	\end{center}
     \vspace{-0.2cm}
	\captionsetup{font={small}}\caption{The principle of hierarchical segmentation, $H_0\subseteq H_1\subseteq\cdots\subseteq H_m$.}
\end{figure}

\subsection{The Convergence Property of AMR}
By comparing Fig. 6 with Fig. 8, it can be observed that AMR provides significant gradient images and AMR-WT generates convergent segmentation results via enlarging the scale of structuring elements. An important convergence property of AMR is described in the following.

\textbf{Theorem 2.} Let $\psi$ be an adaptive morphological reconstruction operator, $\psi$ is convergent when increasing the scale parameter $m$, i.e., for any gradient images $f$ and $g$ such that $b_{s}\subseteq\cdots b_{i}\subseteq b_{i+1}\cdots\subseteq b_{m}$, if $min(\psi(g,s,m))\geq max(R_g^{\phi}(f)_{b_{m+1}})$ then
\begin{equation}\label{eq.13}
  \psi(g,s,m)=\psi(g,s,m+j),
\end{equation}
i.e., $\vee_{s\leq i \leq m}\left\{R_g^{\phi}(f)_{b_i}\right\}=\vee_{s\leq i\leq m+j}\left\{R_g^{\phi}(f)_{b_i}\right\}, 1\leq s \leq m, j\in N^+$, and the proof is presented in Appendix C.

According to Fig 9, it can be seen that the gradient result and the corresponding seed image will remain unchanged when $m\geq 12$. This empirically illustrates that the gradient image reconstructed by AMR is convergent when increasing the value of $m$. Besides, the large gradient magnitude is unchanged while the small gradient magnitude converges to ones larger than itself for AMR. However, the large gradient magnitude converges to one smaller than itself while the small gradient magnitude converges to one larger than itself for MR when the structuring element is small. With the increase of the value of $m$, the value of gradient magnitudes finally converges to the minimum of the original gradient image, i.e., $\lim\limits_{m\rightarrow\infty}R_g^{\phi}(f)_{b_m}=min(g)$ (see Appendix A). Consequently, MR removes all regional minima while AMR only filters useless regional minima and preserves significant ones when $m\rightarrow\infty$.

Furthermore, we analyze how to determine the parameter $m$ for AMR. The computational efficiency of AMR is influenced by the parameter $m$. A small $m$ means a low computational complexity. According to Theorem 2, the reconstructed gradient image and the corresponding segmentation result are unchanged when $min(\psi(g,s,m))\geq max(R_g^{\phi}(f)_{b_{m+1}})$, but the obtained $m$ is usually large. As the paper aims at employing AMR to improve seeded segmentation algorithms, we replace the convergence condition $min(\psi(g,s,m))\geq max(R_g^\phi(f)_{b_{m+1}})$ with checking the difference between $\psi(g,s,m)$ and $\psi(g,s,m-1)$. We propose an objective function for justifying the convergence of AMR
\begin{equation}\label{eq.14}
  J(g,s)=max|\psi(g,s,m)-\psi(g,s,m-1)|,
\end{equation}
where $m\geq2$, $m\in N^+$. It is clear that the segmentation result will remain unchanged when $J\leq\eta$, $\eta$ is a minimal threshold error, and it is a constant used for $J$, but $m$ is a variant for $\psi(g,s,m)$. Consequently, only a parameter $s$ needs to be tuned for obtaining different reconstruction results.

\begin{figure}[t]\label{fig.12}
\renewcommand{\captionlabeldelim}{.}\small
	    \begin{center}
	        \includegraphics[width=1.0\linewidth]{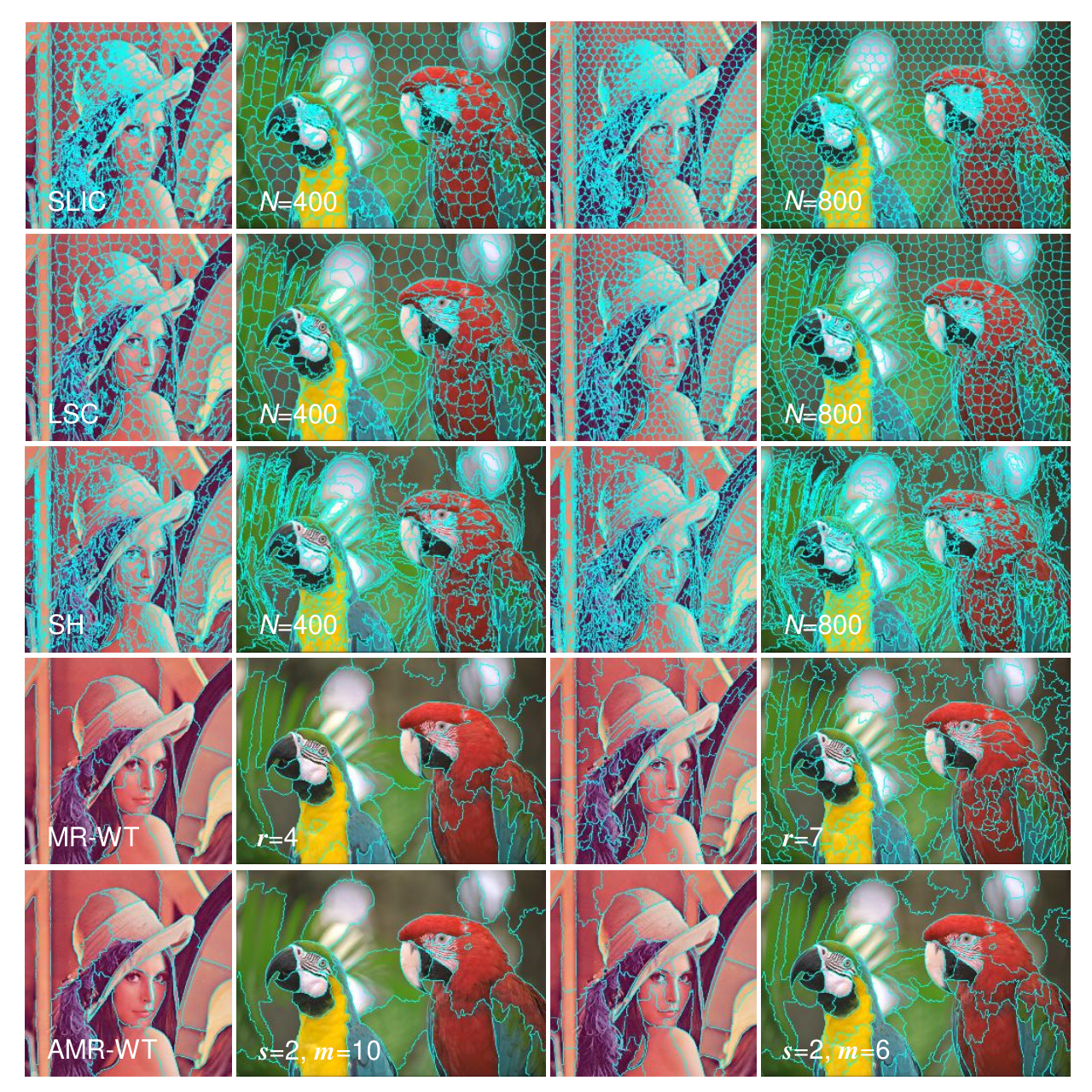}
    	\end{center}
     \vspace{-0.1cm}  %
	\captionsetup{font={small}}\caption{Segmentation results using SLIC, LSC, SH, MR-WT, and AMR-WT, respectively on images with complex texture.Here, $N$ denotes the number of superpixel areas; $N$ is 400 for the left two images and $N$ is 800 for the right two images. $r$ is radius of structuring elements for MR-WT, values of $r$ are $7$ and $4$, respectively. For AMR-WT, $s=2$ and $m=10$ are used for the left two images, while $s=2$ and $m=6$ are used for the right two images.}
\end{figure}

\begin{figure}[t]\label{fig.13}
\renewcommand{\captionlabeldelim}{.}\small
	    \begin{center}
	        \includegraphics[width=1.0\linewidth]{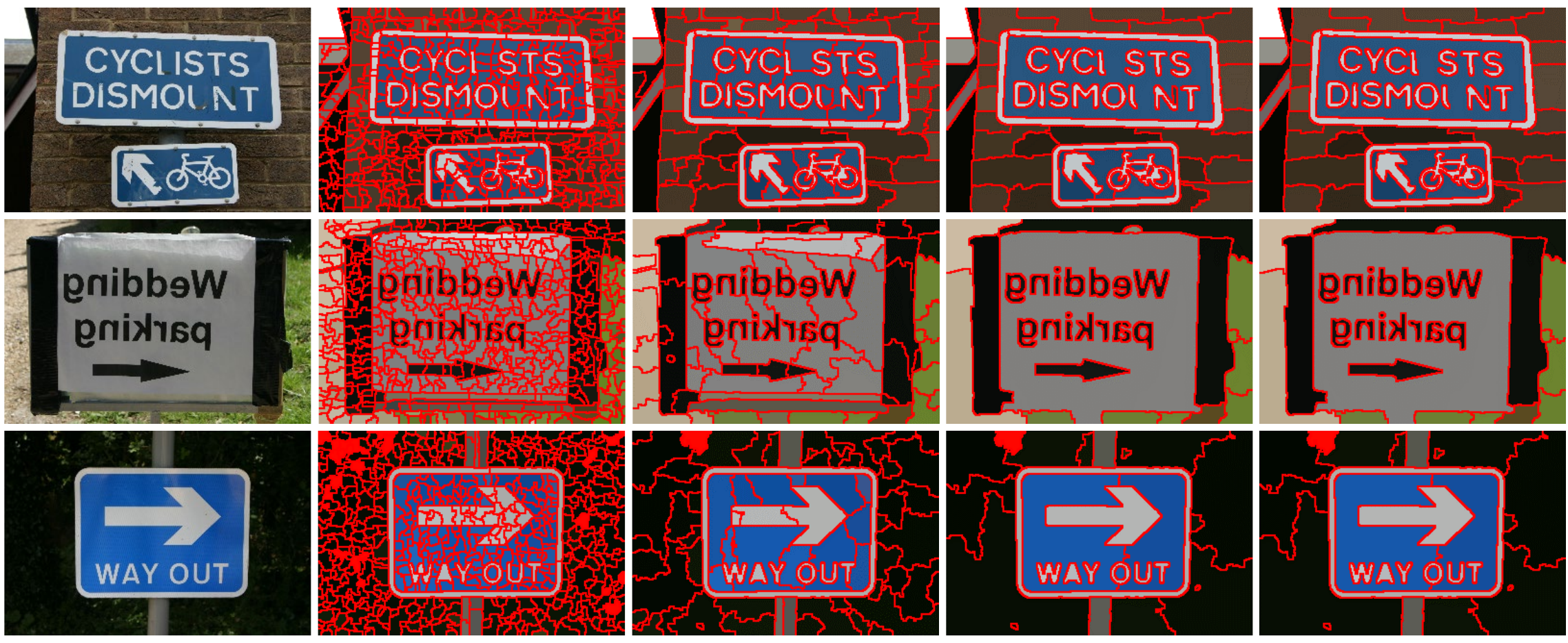}
  {(a)~~~~~~~~~~~~~~(b)~~~~~~~~~~~~(c)~~~~~~~~~~~~~(d)~~~~~~~~~~~~~(e)}
    	\end{center}
     \vspace{-0.1cm}  %
	\captionsetup{font={small}}\caption{Segmentation results using AMR-WT by changing the value of $m$. The results shows that AMR is monotonic increasing by increasing the value of $m$. Moreover, AMR is convergent because the segmentation result is unchanged when $m\geq11$. (a) Images. (b) $s=1$, $m=1$. (c) $s=1$, $m=3$. (d) $s=1$, $m=11$. (e) $s=1$, $m=50$.}
\end{figure}

\begin{algorithm}
  \caption{Adaptive morphological reconstruction (AMR)}

\textbf{Input}:  $g$ (a gradient image).

\textbf{Output}:  $\psi$ (a reconstructed gradient image).

~~1: \textbf{Initialize:} set values for $s$, $m$ (the scale of the minimal and maximal structuring element) and $\eta$, both $m$ and $\eta$  are the convergent condition used for AMR.

~~2:~\textbf{for} $i=s,s+1,\cdots, m$ \textbf{do}

~~3:~Compute $R_{g}^{\phi}(f)_{b_{i}}$ where $f=\varepsilon_{b_{i}}(g)$, $b_{i}$ is a structuring element.

~~4:~Update $\psi(g,s,i)$ and $J(g,s)$,

~~5:~~~\textbf{if} $i=s$ \textbf{then}

~~6:~~~~~$\psi(g,s,i)=\left\{{R_{g}^{\phi}(f)_{b_{i}}}\right\}$

~~7:~~~~~$J(g,s)=max|\psi(g,s,i)|$

~~8:~~~\textbf{else}

~~9:~~~~~~$\psi(g,s,i)=\vee\left\{{\psi(g,s,i-1), R_{g}^{\phi}(f)_{b_{i}}}\right\}$

~~10:~~~~$J(g,s)=max|\psi(g,s,i)-\psi(g,s,i-1)|$

~~11:~~\textbf{end if}



~~12:~~\textbf{if} $J\leq \eta$ \textbf{then}

~~13:~~~~~\textbf{break}

~~14:~~\textbf{end if}

~~15: \textbf{end for}

\end{algorithm}

\subsection{The Algorithm of AMR} 
AMR only involves the parameter $s$ and $\eta$, as described in the detailed steps of AMR in Algorithm 1\footnote{Source code is available at {\color[rgb]{1,0,0.5} https://github.com/SUST-reynole/AMR}}. To speed up the convergence of Algorithm 1, the three parameters $s$, $m$, and $\eta$ are used for AMR because the iteration can be stopped according to $m$ or $\eta$. The computational complexity of AMR depends on the values of $m$ or $\eta$. A large value of $m$ corresponds to a small value of $\eta$. The larger is the value of $m$, the longer is the execution time of AMR. Since we have known that AMR has a fast convergent property as shown in Fig. 8, a small $m$ is enough for moderate-sized images in practical applications. A small m indicates that AMR has a low computational complexity.

Note that the parameter $m$ is unnecessary theoretically, we use two convergent condition $m$ and $\eta$ to speed up the convergence of Algorithm 1. We applied Algorithm 1 to images with complex texture content to demonstrate that the proposed AMR is effective for reducing over-segmentation as shown in Fig. 12.  AMR-WT not only overcomes the problem of over-segmentation but also obtains better contours than MR-WT and state-of-the-art superpixel methods. Furthermore, we test Algorithm 1 on images with text to show the monotonic increasing and convergent properties of AMR. Fig. 13 shows the comparison results. We can see that the segmentation results are nested, which demonstrates the monotonic increasing property of AMR. Moreover, the segmentation results are unchanged when $m\geq 11$, which demonstrates the convergent property of AMR.

\begin{figure}[t]\label{fig.14}
\renewcommand{\captionlabeldelim}{.}\small
	    \begin{center}
	        \includegraphics[width=1.0\linewidth]{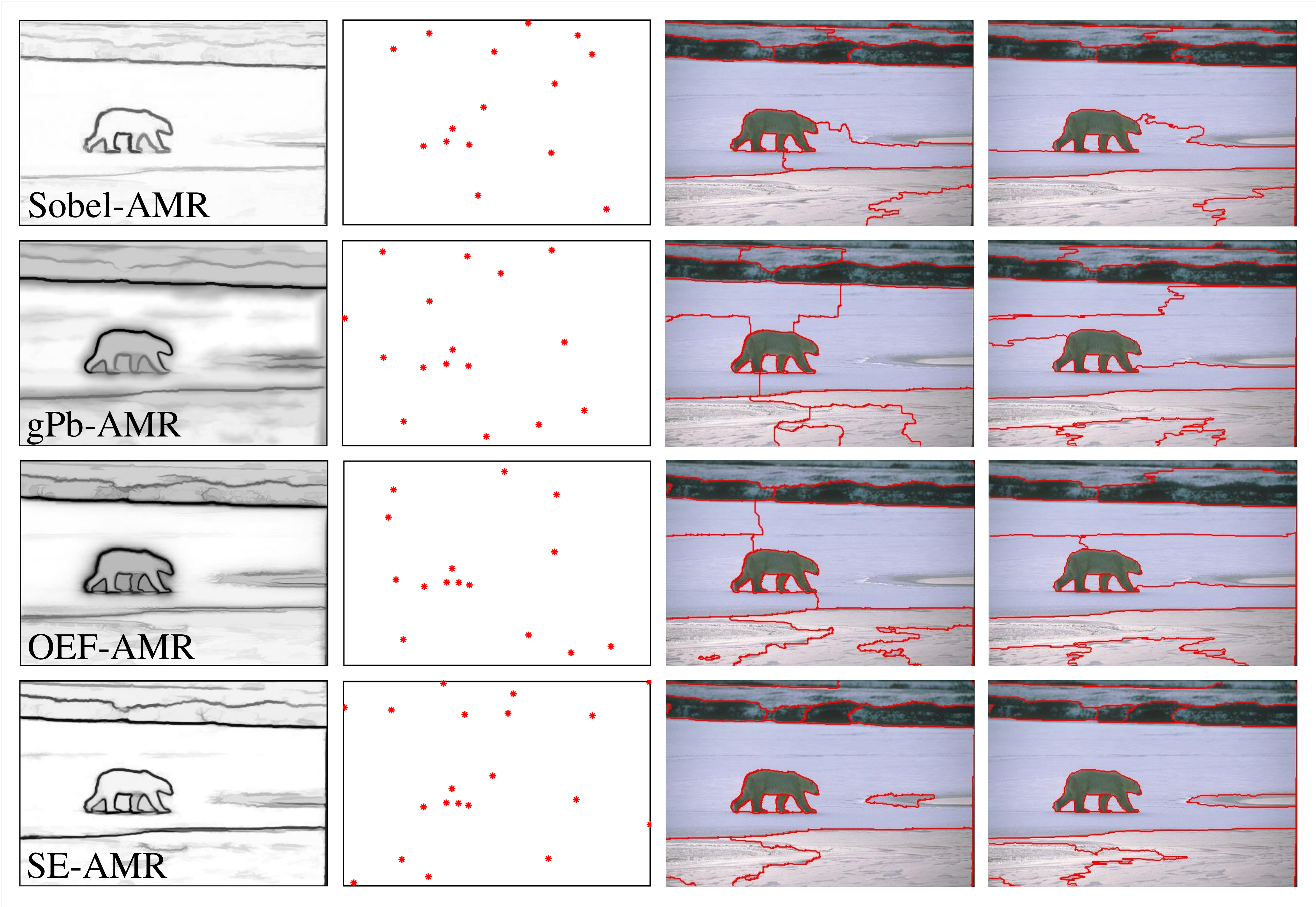}
  {(a)~~~~~~~~~~~~~~~(b)~~~~~~~~~~~~~~~~~(c)~~~~~~~~~~~~~~~~(d)}
    	\end{center}
     \vspace{-0.1cm}  
	\captionsetup{font={small}}\caption{Comparison of segmentation results using AMR-WT/PW ($s=2$). (a) Gradient images. (b) Seeds (regional minimum). (c) WT. (d) PW ($q=2$) [8].}
\end{figure}

\section{Experiments}
To demonstrate the effectiveness and efficiency of the proposed AMR, we apply AMR to seeded image segmentation and spectral segmentation. We conduct experiments on the BSDS500 dataset. The experiments are performed on a workstation with an Intel Core (TM) i7-6700, 3.4GHz CPU and 16GB memory.

We compare the proposed algorithms with state-of-the-art algorithms including a multiscale morphological gradient for watersheds (MMG-WT) [26], multiscale ncut (MNCut) [52], oriented-watershed transform-ultrametric contour map (gPb-owt-ucm) [15], the algorithm recovering occlusion boundaries from an image proposed by Hoiem (gPb-Hoiem) [53], spectral segmentation algorithms proposed by Kim \emph{et al.} (FNCut, cPb-owt-ucm) [39], Higher-order correlation clustering (HO-CC) [54], global/regional affinity graph (GL-graph) [55], single-scale combinatorial grouping (SCG) [49], and multiscale combinatorial grouping (MCG) [49]. The open source codes and model parameters suggested by the corresponding authors are used. Because the author did not present specific parameter values for MMGR-WT, we set $r=5$ and $0.1\leq h\leq 0.3$, where $h$ is a threshold and it is used to generate a marker image, and $r$ is the radius of the structuring element used for MR. For the proposed approaches, we set $1\leq s\leq 3$, $m=50$, and $\eta=10^{-4}$.

We report the experimental results using three evaluation metrics to quantitatively measure the performance of segmentation algorithms: probabilistic rand index (PRI), segmentation covering (CV), and variation of information (VI). The PRI and CV are similarity measures, and they are large while the VI is small when the final segmentation is close to ground truth segmentation.

  \begin{table}[t]\label{table.1}
    \renewcommand{\captionlabeldelim}{.}
    \captionsetup{font={small}}\caption{Comparison of the number of seeds generated by gradient images}
    \vspace{0.1cm}
       \renewcommand{\baselinestretch}{1.25}\footnotesize\centerline{\tabcolsep=6.5pt{\begin{tabular}{p{3cm} c c c c}
       \Xhline{1.2pt}
                Images                                 &Sobel	&gPb [15]     &OEF [17]    &SE [16]\\
       \Xhline{1.2pt}
                Original gradient images	 	                &9175	  &746	  &5348  	&1347\\
                Gradient images reconstructed by AMR        &\multirow{2}{*}{15}    &\multirow{2}{*}{16}    &\multirow{2}{*}{15}    &\multirow{2}{*}{19}\\
      \Xhline{1.2pt}
       \end{tabular}}}
    \end{table}

\subsection{Seeded Image Segmentation}
AMR is useful for improving seeded image segmentation because it employs multiscale structuring elements to obtain a convergent seed image without pre-setting many parameters. To show the capability of AMR, it is applied to different gradient images to filter seeds. Fig. 14 shows reconstructed gradient images by AMR and the corresponding segmentation results by WT/PW. These results are clearly better than the ones shown in Fig. 5. The problem of over-segmentation for seeded segmentation algorithms is therefore addressed. Furthermore, compared Fig. 6 to Fig. 14, although both MR and AMR are able to filter seeds, AMR is able to maintain meaningful seeds that correspond to important contours.

Furthermore, Table I shows the number of seeds generated by gradient images. We can see that the reconstructed gradient images generate fewer seeds than original gradient images, which demonstrates AMR is efficient for the filtering of useless seeds. Moreover, AMR is robust for different gradient images obtained by Sobel, gPb, OEF, and SE because the final segmentation results are similar.

In Fig. 14, we set $s=2$ because the segmentation result includes too many small regions when $s=1$. Clearly, $s$ controls the number of small regions in segmentation results. Generally, the value of $s$ depends on the resolution of the images to be segmented, e.g., $1\leq s\leq 3$ for BSDS500.

To demonstrate that the proposed AMR is robust for different images, we implement AMR-WT/PW on the BSDS500. Fig. 15 shows the comparison of segmentation results using different algorithms, i.e., Sobel-AMR-WT/PW, gPb-AMR-WT/PW, OEF-AMR-WT/PW, and SE-AMR-WT/PW. The segmentation results demonstrate the effectiveness of AMR for the filtering of useless seeds, Moreover, AMR is effective for both WT and PW.

\begin{figure}[t]\label{fig.15}
\renewcommand{\captionlabeldelim}{.}\small
	    \begin{center}
	        \includegraphics[width=1.0\linewidth]{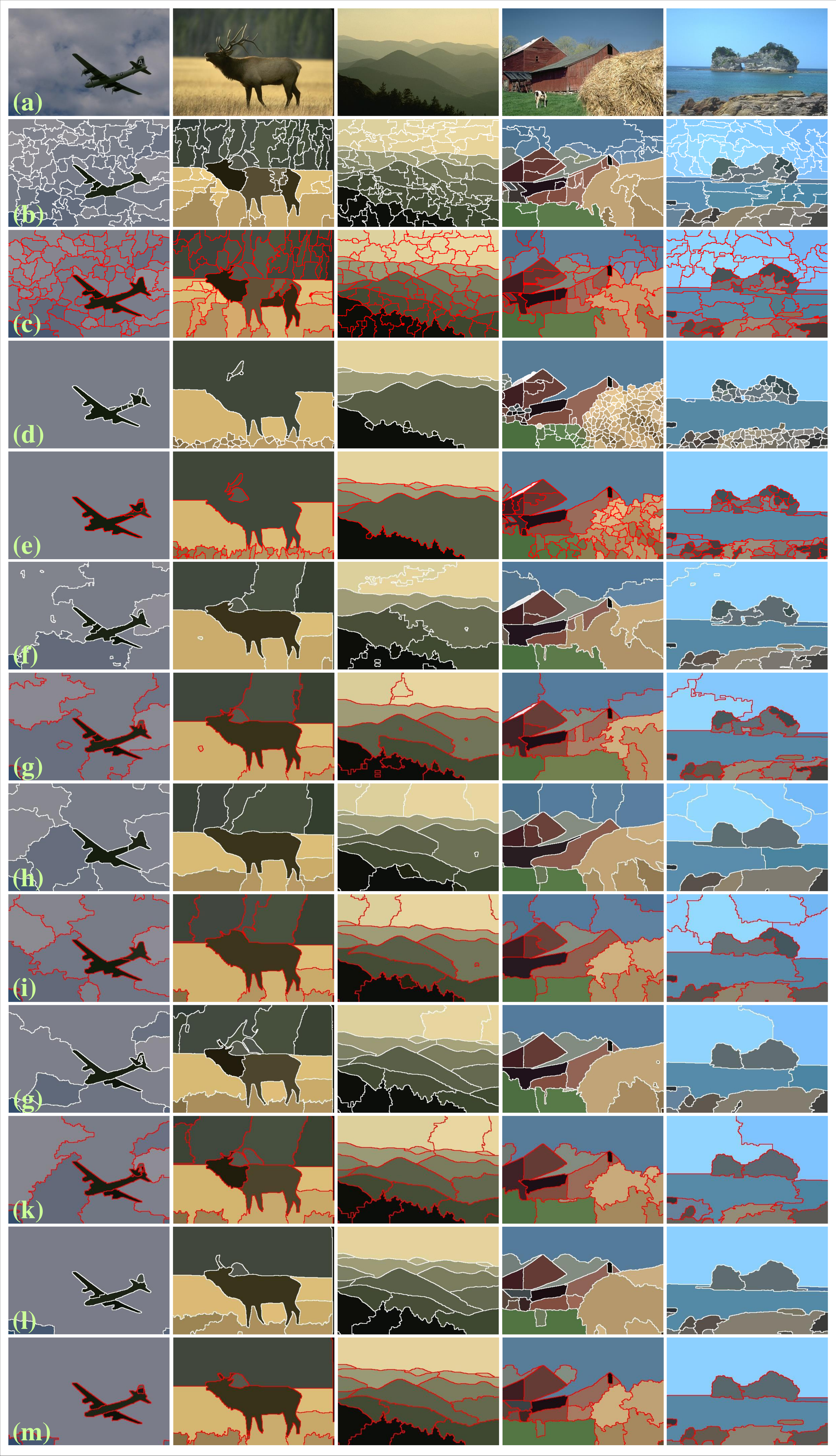}
    	\end{center}
     \vspace{-0.1cm}
	\captionsetup{font={small}}\caption{Comparison of segmentation results using different algorithms ($s=2$). (a) Images. (b) Sobel-MR-WT($r=5$). (c) Sobel-MR-PW ($r=5$). (d) MMG-MR-WT ($r=5$ and $h=0.2$). (e) MMG-MR-PW ($r=5$ and $h=0.2$). (f) Sobel-AMR-WT. (g) Sobel-AMR-PW. (h) gPb-AMR-WT. (i) gPb-AMR-PW. (j) OEF-AMR-WT. (k) OEF-AMR-PW. (l) SE-AMR-WT. (m) SE-AMR-PW.}
\end{figure}

To compare the performance of different algorithms on the BSDS500, Table II shows experimental results of three evaluation metrics: PRI, CV, and VI. We can see that AMR is more efficient for improving segmentation results obtained by WT or PW compared to MR. MR is sensitive to $r$ while AMR is insensitive to $s$. Although MMG-MR-WT/PW is effective for the over-segmentation reduction by introducing the parameter $h$, segmentation results are sensitive to both $r$ and $h$. The gPb-AMR-WT/PW, OEF-AMR-WT/PW, and SE-AMR-WT/PW obtain better performance than Soble-AMR-WT/PW since the former provides better gradient images. The SE-AMR-WT/PW obtains the best performance. In addition, AMR-WT obtains higher PRI, CV, and lower VI than AMR-PW in the same situation.

Because AMR converges quickly, AMR has a high computation efficiency for gradient reconstruction. Table III shows the comparison of running time of AMR-WT on different gradient images obtained by Sobel, gPb, OEF, and SE, respectively. We only present the running time of AMR-WT here because AMR-PW has a similar running time as AMR-WT. It can be seen from Table III that AMR-WT has a short running time to achieve image segmentation on the BSDS500. The SE-AMR-WT requires the shortest running time because the corresponding gradient image converges quicker under AMR. Tables II-III show AMR is effective and efficient for improving seeded segmentation algorithms such as WT and PW.

Additional evidence of the superiority of AMR can be found in Fig. 16 which shows experimental results on images with rich texture and faded boundaries. According to Figs. 15-16, we can see that the proposed AMR is effective for different kinds of images.

\begin{table}[t]\label{table.2}
  \renewcommand{\captionlabeldelim}{.}
  \captionsetup{font={small}}\caption{Quantitative results (PRI, CV, VOI) on the BSDS500. Larger is better for PRI and CV while smaller is better for VI. The best values are in bold.}
    \vspace{0.1cm}
      \renewcommand{\baselinestretch}{1.25}
      {\footnotesize\centerline{\tabcolsep=6pt\begin{tabular}{l c c c}
      \Xhline{1.2pt}
        Methods                         	    &PRI$\uparrow$	&CV$\uparrow$     &VI$\downarrow$\\
      \Xhline{1.2pt}
              Sobel-MR-WT/PW  ($r=5$)	         	        &0.71/0.71	&0.16/0.15	&4.02/4.12	\\
              Sobel-MR-WT/PW  ($r=8$)	 	                &0.73/0.73	&0.28/0.28	&3.08/3.05	\\
              Sobel-MR-WT/PW  ($r=12$)	      	          &0.69/0.68	&0.38/0.39	&2.67/2.33	\\
              MMG-MR-WT/PW [26]  ($h=0.1$)	 	                &0.76/0.76	&0.27/0.27	&4.47/4.45	\\
              MMG-MR-WT/PW [26]  ($h=0.2$)	 	                &0.74/0.74	&0.38/0.38	&3.50/3.45	\\
              MMG-MR-WT/PW [26]  ($h=0.3$)	 	                &0.62/0.62	&0.42/0.42	&2.95/2.92	\\
              Sobel-AMR-WT/PW ($s=1$)	 	                &0.76/0.75	&0.39/0.34	&2.54/2.79	\\
              Sobel-AMR-WT/PW ($s=2$)	         	        &0.76/0.75	&0.39/0.36	&2.51/2.70	\\
              Sobel-AMR-WT/PW ($s=3$)		                &0.76/0.76	&0.39/0.36	&2.52/2.66	\\
              gPb-AMR-WT/PW ($s=1$)               &0.75/0.75	&0.35/0.32	&2.55/2.77	\\
              gPb-AMR-WT/PW ($s=2$)	    	        &0.75/0.74	&0.35/0.33	&2.55/2.77	\\
              gPb-AMR-WT/PW ($s=3$)	   	          &0.75/0.74	&0.36/0.33	&2.55/2.76	\\
              OEF-AMR-WT/PW ($s=1$)              &0.77/0.75	&0.39/0.34	&2.45/2.72	\\
              OEF-AMR-WT/PW ($s=2$)              &0.77/0.75	&0.39/0.34	&2.43/2.72	\\
              OEF-AMR-WT/PW ($s=3$)              &0.77/0.76	&0.39/0.35	&2.41/2.70	\\
              SE-AMR-WT/PW ($s=1$)               &\textbf{0.80}/0.79	&\textbf{0.45}/0.41	&2.25/2.52	\\
              SE-AMR-WT/PW ($s=2$)               &\textbf{0.80}/0.79	&\textbf{0.45}/0.41	&2.23/2.51	\\
              SE-AMR-WT/PW ($s=3$)               &\textbf{0.80}/0.79	&\textbf{0.45}/0.41	&\textbf{2.21}/2.50	\\
    \Xhline{1.2pt}
    \end{tabular}}}
  \end{table}

\begin{table}[t]\label{table.3}
  \renewcommand{\captionlabeldelim}{.}
  \captionsetup{font={small}}\caption{Comparison of average running time of AMR-WT on the BSDS500 (in seconds). Lower is better. The best values are in bold ($\eta=10^{-4}$).}
  \vspace{0.1cm}
      \renewcommand{\baselinestretch}{1.25}
      {\footnotesize\centerline{\tabcolsep=16pt\begin{tabular}{l c c c c}
      \Xhline{1.2pt}
              $s$                                 &Sobel	&gPb     &OEF    &SE\\
      \Xhline{1.2pt}
              $s=1$	 	     &0.801	&1.278	&0.861	&\textbf{0.565}\\
              $s=2$        &0.774	&1.241	&0.825	&\textbf{0.534}\\
              $s=3$        &0.689	&1.169	&0.766	&\textbf{0.480}\\
    \Xhline{1.2pt}
      \end{tabular}}}
  \end{table}

\begin{figure}[t]\label{fig.16}
\renewcommand{\captionlabeldelim}{.}\small
	    \begin{center}
	        \includegraphics[width=1.0\linewidth]{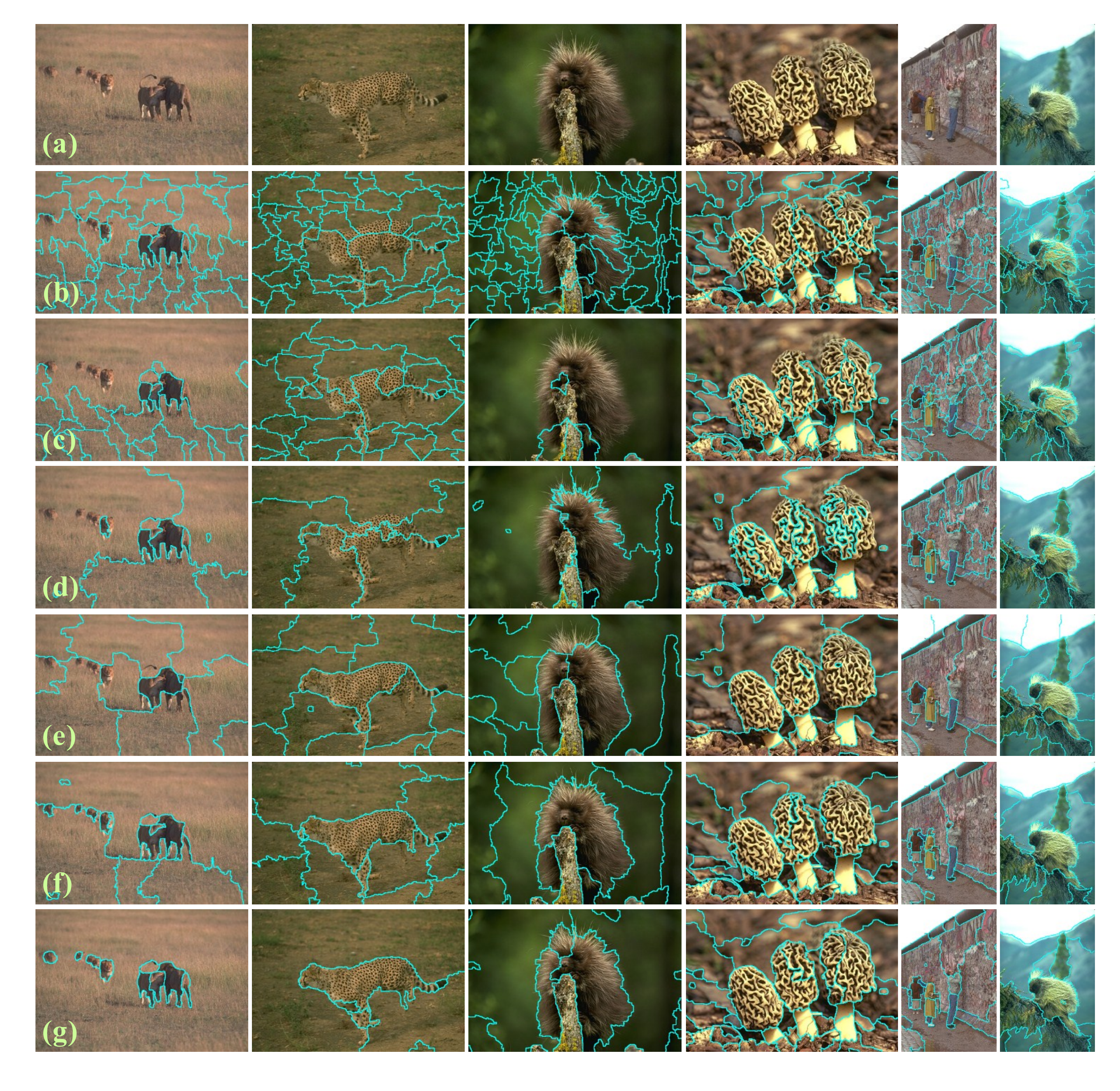}
    	\end{center}
     \vspace{-0.1cm}  
	\captionsetup{font={small}}\caption{Comparison of segmentation results using different algorithms ($s=2$). (a) Images with rich texture or faded boundaries. (b) Sobel-MR-WT ($r=5$). (c) MMG-MR-WT ($r=5$ and $h=0.2$). (d) Sobel-AMR-WT. (e) gPb-AMR-WT. (f) OEF-AMR-WT. (g) SE-AMR-WT.}
\end{figure}

\subsection{Seed-based Spectral Segmentation}
In this section, we directly construct the affinity matrix on a pre-segmentation image provided by AMR-WT to reduce the size of the affinity matrix, and then compute the subsequent steps of spectral segmentation (we name it AMR-SC). Note that we employ AMR-WT rather than AMR-PW because the former is able to provide better pre-segmentation results than the latter as shown in Table II. As the pre-segmentation image only consists of dozens of regions, we consider color feature in CIELAB color space and Gaussian function as the criterion to measure the similarity of two regions. Throughout the paper, we use $\sigma=1$. It is clear that the affinity matrix produced by AMR is a small matrix. Therefore, the clusters can be detected easily and fast with the $k$-means algorithm.

\begin{figure*}[t]\label{fig.17}
\renewcommand{\captionlabeldelim}{.}\small
	    \begin{center}
	        \includegraphics[width=1.0\linewidth]{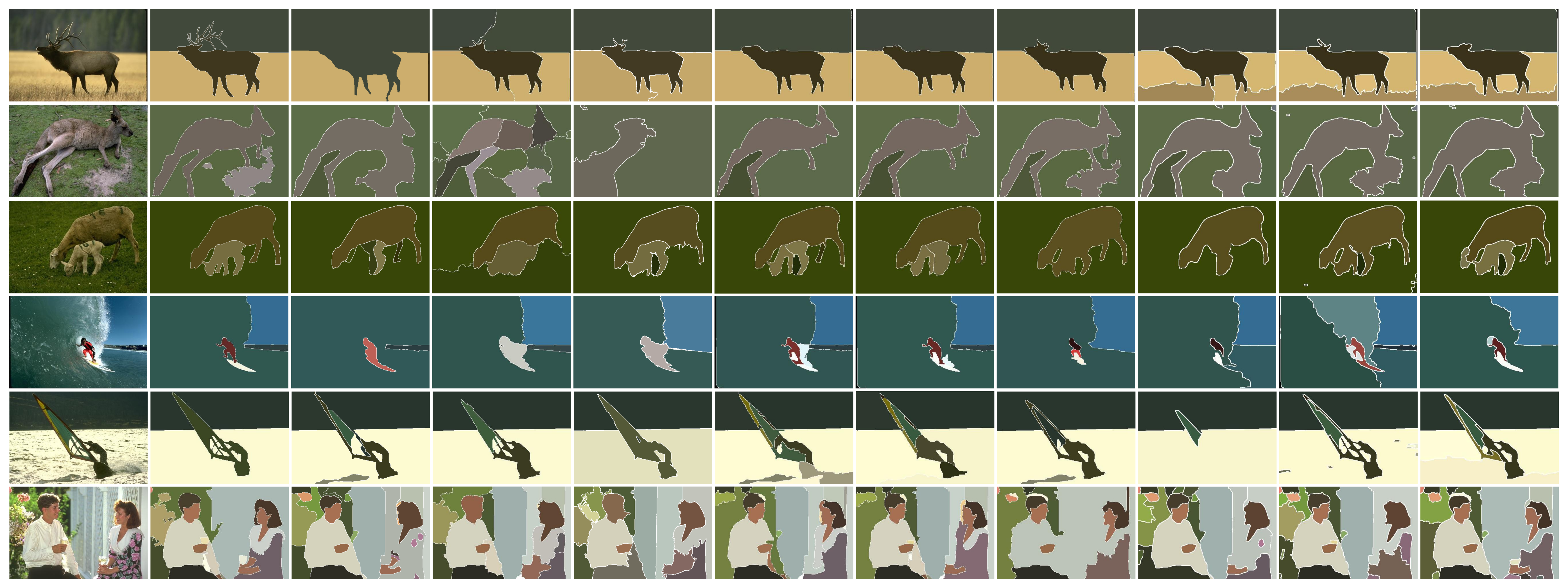}
{(a)~~~~~~~~~~~~(b)~~~~~~~~~~~~(c)~~~~~~~~~~~~(d)~~~~~~~~~~~~(e)~~~~~~~~~~~~(f)~~~~~~~~~~~~(g)~~~~~~~~~~~~(h)~~~~~~~~~~~~(i)~~~~~~~~~~~~(j)~~~~~~~~~~~~(k)}
    	\end{center}
     \vspace{-0.1cm}  
	\captionsetup{font={small}}\caption{Comparison of segmentation results on the BSDS500 using different algorithms ($s=2$). (a) Images. (b) Ground truths. (c) gPb-owt-ucm. (d) FNCut. (e) GL-graph. (f) SCG. (g) MCG. (h) Sobel-AMR-SC. (i) gPb-AMR-SC. (j) OEF-AMR-SC. (k) SE-AMR-SC.}
\end{figure*}

\begin{figure}[t]\label{fig.18}
\renewcommand{\captionlabeldelim}{.}\small
	    \begin{center}
	        \includegraphics[width=1.0\linewidth]{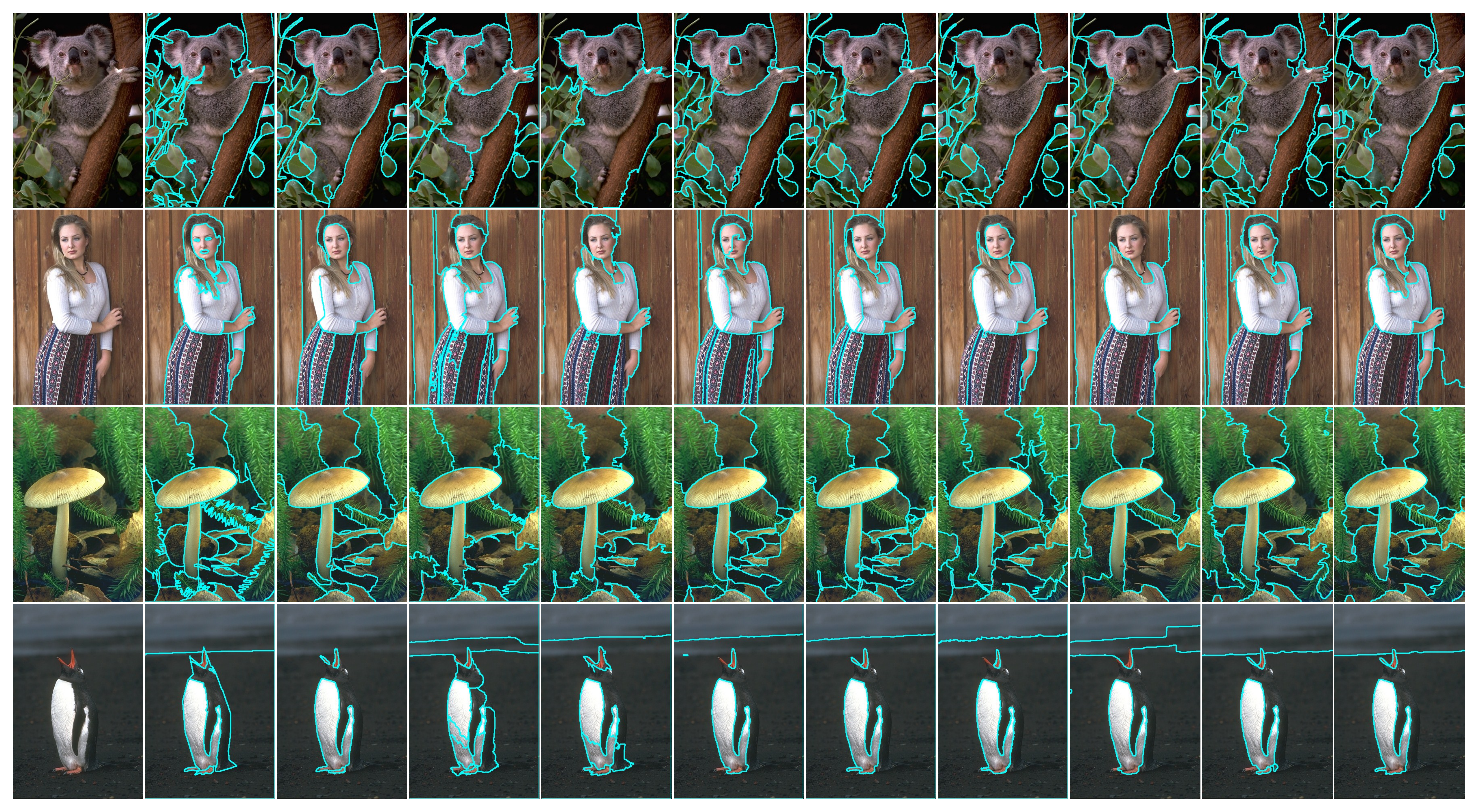}
 {(a)~~~~(b)~~~~(c)~~~~(d)~~~~(e)~~~~(f)~~~~(g)~~~~(h)~~~~(i)~~~~(j)~~~~(k)}
    	\end{center}
     \vspace{-0.1cm}  
	\captionsetup{font={small}}\caption{Comparison of segmentation results on the BSDS500 using different algorithms ($s=2$). Compared to Figs. 15 and 17, we use an overlay of the segmented result with respect to the original image to show the accuracy of the boundary. (a) Images. (b) Ground truths. (c) gPb-owt-ucm. (d) FNCut. (e) GL-graph. (f) SCG. (g) MCG. (h) Sobel-AMR-SC. (i) gPb-AMR-SC. (j) OEF-AMR-SC. (k) SE-AMR-SC.}
\end{figure}

In this paper, the pre-segmentation depends on AMR. According to Table II, we set $s=2$ and $3$, and we set the number of clusters for the $k$-means according to [39], [55]. The proposed AMR-SC is evaluated on BSDS500 and compared with algorithms such as gPb-owt-ucm, FNCut, GL-graph, SCG and MCG. Figs. 17-18 show that the proposed AMR-SC generates better segmentation results than those comparative algorithms. The result demonstrates that AMR is useful for improving spectral segmentation due to two reasons. The first is that the regional spatial information of an image provided by pre-segmentation is integrated into spectral segmentation, and the second is that the affinity graph is reduced efficiently by removing useless seeds.

Furthermore, we employ the three measures: PRI, CV and VI to compare the proposed AMR-SC with nine state-of-the-art image segmentation algorithms. Table VI shows the region benchmarks on the BSDS500. In Table VI, the proposed AMR-SC clearly dominates other algorithms on PRI and CV, and is on par with SCG on VI mainly due to accurate pre-segmentation provided by AMR-WT. The OEF-AMR-SC and SE-AMR-SC provide better performance than gPb-AMR-SC and Sobel-AMR-SC because OEF and SE obtain better gradient images than gPb and Sobel. In addition, AMR-SC is insensitive to the parameter $s$.

We tested the running time complexity on the BSDS500 dataset. The running time comparison is shown in Table IV. On average, generating a pre-segmentation result with SE-AMR-WT takes 0.54 seconds (SE generates a gradient image requiring 0.06 seconds. AMR-WT takes 0.48 seconds, $s=3$ and $\eta=10^{-4}$), and constructing an affinity graph and spectral clustering take 0.059 seconds. Consequently, SE-AMR-SC takes about 0.60 second to segment an image from the BSDS500. In contrast, the gPb-owt-ucm takes almost 106.38 seconds, FNCut takes about 10.58 seconds. As GL-graph has four steps, i.e., over-segmentation, feature extraction, bipartite graph construction and graph partition using spectral clustering, it is more complex than SE-AMR-SC, and takes almost 7.41 seconds. MCG takes about 18.60s per image to compute the multiscale hierarchy but SCG takes only 2.21s per image. It is clear that our SE-AMR-SC is the fastest because AMR-SC only depends on the gradient information, and the generated affinity matrix is small.

\subsection{Discussion}
AMR has two parameters, $\eta$ and $s$. $\eta$ relates to the convergent condition. Generally, a large value of $\eta$ means a few iterations (a small $m$, where $m$ is the number of iterations) while a small value of $\eta$ corresponds to many iterations (a large $m$). Table V shows the influence of $\eta$ on $m$ for test images. We can see that $m$ increases with the decrease of $\eta$ but $m$ is unchanged when $\eta\leq 10^{-4}$.

Furthermore, to show the influence of $\eta$ on AMR, we implement AMR on the BSDS500 by setting different values of $\eta$, and Tables VI-VII show the results. It is clear that the number of iterations for AMR-WT is smaller and running time is shorter if the value of $\eta$ is larger. However, the number of iterations and running time are unchanged when $\eta\leq 10^{-4}$. Therefore, in practical application, users can select different values of $\eta$ according to their requirements.

\begin{table}[t]\label{table.4}
  \renewcommand{\captionlabeldelim}{.}
  \captionsetup{font={small}}\caption{Quantitative results (PRI, CV, and VI) on the BSDS500. Larger is better for PRI and CV while smaller is better for VI and running time. The best values are in bold.}
   \vspace{0.1cm}

     \renewcommand{\baselinestretch}{1.25}
     {\footnotesize\centerline{\tabcolsep=11pt\begin{tabular}{lcccc}
     \Xhline{1.0pt}
              Methods                   	            &PRI$\uparrow$	&CV$\uparrow$	&VI$\downarrow$ &Time$\downarrow$\\
     \Xhline{1.0pt}
              MNCut [52]	                          &0.78	&0.45	&2.23  &37.25\\
              gPb-owt-ucm [15]	                    &0.83	&0.59	&1.69  &106.38\\		
              gPb-Hoiem [53]	                      &0.81	&0.56	&1.78  &109.77\\				
              FNCut [39]	                          &0.81	&0.53	&1.86 &10.58\\		
              cPb-owt-ucm [39]	                    &0.83	&0.59	&1.65 &107.13\\
              HO-CC [54]	                          &0.83	&0.60	&1.79  &35.18\\		
              GL-graph [55]                     &0.84 &0.59 &1.80 &7.41\\
              SCG [49]                            &0.83	&0.60	&1.63  &2.21\\		
              MCG [49]	                            &0.83	&0.61	&\textbf{1.57} &18.60\\		
              Sobel-AMR-SC ($s=2$)	                &0.82	&0.61	&1.77 &0.86\\		
              Sobel-AMR-SC ($s=3$)	                &0.82	&0.61	&1.77 &0.81\\
              gPb-AMR-SC ($s=2$)                    &0.82	&0.61	&1.73 &102.94\\
              gPb-AMR-SC ($s=3$)                    &0.82	&0.61	&1.73 &102.92\\
              OEF-AMR-SC ($s=2$)	                    &\textbf{0.85}	&\textbf{0.63}	&1.62 &6.16\\
              OEF-AMR-SC ($s=3$)	                  &0.84	&\textbf{0.63}	&1.64  &6.07\\
              SE-AMR-SC ($s=2$)	                    &\textbf{0.85}	&\textbf{0.63}	&1.62  &0.62\\
              SE-AMR-SC ($s=3$)	                    &\textbf{0.85}	&\textbf{0.63}	&1.62  &\textbf{0.60}\\
    \Xhline{1.0pt}
     \end{tabular}}}
  \end{table}

  \begin{table}[t]\label{table.5}
  \renewcommand{\captionlabeldelim}{.}
  \captionsetup{font={small}}\caption{The number of iterations of SE-AMR-WT under different values of $\eta$, ($s=3$). The number of iterations $m$ is unchanged when $\eta\leq 10^{-4}$, and the invariant values of $m$ are in bold.}
   \vspace{0.1cm}

     \renewcommand{\baselinestretch}{1.25}
     {\footnotesize\centerline{\tabcolsep=8.5pt\begin{tabular}{l c c c c}
     \Xhline{1.2pt}
              Images                   	            &$\eta=10^{-2}$	&$\eta=10^{-3}$	&$\eta=10^{-4}$ &$\eta=10^{-5}$\\
     \Xhline{1.2pt}
     ``2092"  & 9  & 16 &\textbf{16} &\textbf{16} \\
     ``8023"  & 6 & 12 &\textbf{19} &\textbf{19} \\
     ``8049"  & 13 & 16 &\textbf{19} &\textbf{19} \\
     ``12074" & 9 & 17 &\textbf{17} &\textbf{17} \\
     ``12084" & 9 & 13 &\textbf{13} &\textbf{13} \\
     ``15004" & 11 & 21 &\textbf{21} &\textbf{21} \\
    \Xhline{1.2pt}
     \end{tabular}}}
    \end{table}

Furthermore, we implemented SE-AMR-WT on BSDS500 with different values of $\eta$. The performance indices of segmentations are shown in Table VIII. By comparing Tables V-VIII, we can see that the average number of iterations, running time, and segmentation accuracy are unchanged for AMR-WT when $\eta\leq 10^{-4}$. Therefore, the proposed AMR is insensitive to $\eta$.

The value of $s$ controls the initial gradient value of images. A large $s$ will cause the contour offset while a small value of $s$ will cause too many unexpected small regions. Therefore, we choose $s=2$ and $s=3$ for the BSDS500 in Table IV. To further show the influence of $s$ on AMR, Table IX shows the performance indices of segmentations on BSDS500 by setting different values of $s$. It can be seen from Table IX that SE-AMR-WT is insensitive to $s$ if $1\leq s\leq 6$.

\section{Conclusion}
In this work, we have studied the advantages and disadvantages of MR on seeded segmentation algorithms. We proposed an efficient AMR algorithm that can preferably improve seeded segmentation algorithms. The proposed AMR has two significant properties, the monotonic increasingness and the convergence. The monotonic increasingness helps AMR to achieve a hierarchical segmentation. The convergence is able to alleviate the drawback of MR for the filtering of useless regional minima in a gradient image, and guarantees a convergent result. Moreover, we have explored the applications of AMR and have found that AMR is not only able to improve seeded image segmentation results, but also can obtain better spectral segmentation results than state-of-the-art algorithms. Furthermore, the proposed AMR-SC is computationally efficient because a small affinity matrix is used for spectral clustering. Experimental results clearly demonstrate that the proposed AMR-WT generates satisfactory and convergent segmentation results without hard-tuning parameters, and the AMR-SC outperforms most of the state-of-the-art algorithms for image segmentation, and it performs the best in two metrics: PRI and CV.

\begin{table}[t]\label{table.6}
  \renewcommand{\captionlabeldelim}{.}
  \captionsetup{font={small}}\caption{The average number of iterations of SE-AMR-WT under different values of $\eta$, $s=3$. The average number of iterations is unchanged when $\eta\leq 10^{-4}$, and the invariant values of $m$ are in bold.}
   \vspace{0.1cm}
    \renewcommand{\baselinestretch}{1.25}
    {\footnotesize\centerline{\tabcolsep=6pt\begin{tabular}{l c c c c c}
    \Xhline{1.2pt}
              {}                  	  &$\eta=10^{-1}$ &$\eta=10^{-2}$	&$\eta=10^{-3}$	&$\eta=10^{-4}$ &$\eta=10^{-5}$\\
    \Xhline{1.2pt}
    $m$  & 2.0 & 9.3 & 17.0 &\textbf{18.9} & \textbf{18.9}\\
    \Xhline{1.2pt}
    \end{tabular}}}
  \end{table}

\begin{table}[t]\label{table.7}
  \renewcommand{\captionlabeldelim}{.}
  \captionsetup{font={small}}\caption{The average running time of SE-AMR-WT on the BSDS500 (in seconds), $s=3$. The average running time is unchanged when $\eta\leq 10^{-4}$, and the invariant values are in bold.}
    \vspace{0.1cm}
    \renewcommand{\baselinestretch}{1.25}
    {\footnotesize\centerline{\tabcolsep=5pt\begin{tabular}{l c c c c c}
    \Xhline{1.2pt}
              {}                  	  &$\eta=10^{-1}$ &$\eta=10^{-2}$	&$\eta=10^{-3}$	&$\eta=10^{-4}$ &$\eta=10^{-5}$\\
    \Xhline{1.2pt}
    Time  &0.089 &0.228 &0.430 & \textbf{0.480} &\textbf{0.480}\\
    \Xhline{1.2pt}
    \end{tabular}}}
  \end{table}

\begin{table}[t]\label{table.8}
\renewcommand{\captionlabeldelim}{.}
\captionsetup{font={small}}\caption{Quantitative results (PRI, CV and VI) of SE-AMR-WT on the BSDS500 under different values of $\eta$, $s=3$. Higher is better for PRI and CV while lower is better for VI.}
  \vspace{+0.3cm}

    \renewcommand{\baselinestretch}{1.25}
    {\footnotesize\centerline{\tabcolsep=19pt\begin{tabular}
    {l c c c }  
    \Xhline{1.2pt}
            $\eta$  &PRI$\uparrow$	&CV$\uparrow$	&VI$\downarrow$\\
    \Xhline{1.2pt}
    $10^{-1}$  &0.77 & 0.30 &3.21\\
    $10^{-2}$  & 0.79 & 0.39 & 2.52 \\
    $10^{-3}$  & 0.80 & 0.45 & 2.23 \\
    $10^{-4}$  & 0.80 & 0.45 & 2.21 \\
    $10^{-5}$  & 0.80 & 0.45 & 2.21 \\
  \Xhline{1.2pt}
    \end{tabular}}}
\end{table}

The segmentation results generated by AMR-WT or AMR-SC can be directly used in object recognition and scene labeling. However, AMR-WT or AMR-SC cannot obtain semantic segmentation results compared to the popular convolutional neural network (CNN), e.g., fully convolutional network (FCN) [56]. To further improve the contour quality of segmentation results, traditional algorithms such as conditional random field [57], image superpixel [58], and spatial pyramid pooling [59], are used to improve the performance of CNN on image segmentation. AMR can be also used in CNN to improve semantic segmentation results. For our future work, we plan to investigate how to combine AMR and FCN effectively and efficiently.

\begin{table}[t]\label{table.9}
\renewcommand{\captionlabeldelim}{.}
\captionsetup{font={small}}\caption{Quantitative results (PRI, CV and VI) of SE-AMR-WT on the BSDS500 under different values of $s$, $\eta=10^{-4}$. Larger is better for PRI and CV while smaller is better for VI.}
  \vspace{+0.4cm}

    \renewcommand{\baselinestretch}{1.25}
    {\footnotesize\centerline{\tabcolsep=19pt\begin{tabular}
    {l c c c }
    \Xhline{1.2pt}
            $s$  &PRI$\uparrow$	&CV$\uparrow$	&VI$\downarrow$\\
    \Xhline{1.2pt}
    1  &0.80  &0.45  &2.25  \\
    2  &0.80  &0.45  &2.23  \\
    3  &0.80  &0.46  &2.21 \\
    4  &0.80  &0.46  &2.22 \\
    5  &0.80  &0.46  &2.25 \\
    6  &0.79  &0.46  &2.30 \\
  \Xhline{1.2pt}
    \end{tabular}}}
\end{table}

\appendices
\section{Proof of $\lim\limits_{m\rightarrow\infty}R_{g}^{\gamma}(f)_{b_{m}}=max(g)$}
\noindent\textbf{Proof}:

Since
\[f = {\delta _{{b_m}}}\left( g \right),{\rm{ }}m \to \infty ,{\rm{ }}\mathop {\lim }\limits_{m \to \infty } {\delta _{{b_m}}}\left( g \right) = \max \left( g \right)\]

\noindent we have,
\begin{equation*}
f=max(g),
\end{equation*}
\noindent and
\begin{equation*}
\begin{split}
\varepsilon_{g}^{(1)}(f)&=\varepsilon(f)\vee g\\
&=\varepsilon(max(g))\vee g\\
&=max(g),\\
\end{split}
\end{equation*}

\begin{equation*}
\begin{split}
\varepsilon_{g}^{(n)}(f)&=\varepsilon(\varepsilon_{g}^{(n-1)}(f))\vee g\\
&=max(g).\\
\end{split}
\end{equation*}

\noindent According to $R_{g}^{\gamma}(f)=R_{g}^{\delta}(R_{g}^{\varepsilon}(f))$, $R_{g}^{\varepsilon}(f)=\varepsilon_{g}^{n}(f)$ in (1), we get
\begin{equation*}
\begin{split}
R_{g}^{(\varepsilon)}(f)=max(g).
\end{split}
\end{equation*}
\noindent Thus,
\begin{equation*}
\begin{split}
\lim\limits_{m\rightarrow\infty}R_{g}^{\gamma}(f)_{b_{m}}&=\lim\limits_{m\rightarrow\infty}R_{g}^{\delta}(R_{g}^{\varepsilon}(f))\\
&=R_{g}^{\delta}(max(g))\\
&=max(g).\\
\end{split}
\end{equation*}

\noindent In terms of the duality of morphological operation,
\begin{equation*}
\begin{split}
\lim\limits_{m\rightarrow\infty}R_{g}^{\phi}(f)_{b_{m}}=min(g).
\end{split}
\end{equation*}
$\hfill\blacksquare$

\section{Proof of theorem 1}
\begin{equation*}
  p\leq q\Rightarrow\psi(g,s,p)\leq\psi(g,s,q)
\end{equation*}
\noindent\textbf{Proof:}

\noindent Let $s\leq p\leq q\leq m$, from Definition 1, we have
\begin{equation*}
\psi(g,s,p)=\vee\left\{R_g^\phi(f)_{b_s}, R_g^\phi(f)_{b_{s+1}},\cdots, R_g^\phi(f)_{p}\right\},\\
\end{equation*}
\begin{equation*}
\psi(g,s,p)=\vee\left\{R_g^\phi(f)_{b_s}, R_g^\phi(f)_{b_{s+1}},\cdots, R_g^\phi(f)_{q}\right\}.\\
\end{equation*}
Because $p\leq q$,
\begin{equation*}
\psi(g,s,p)=\vee\left\{\psi(g,s,p), R_g^\phi(f)_{b_{p+1}}, \cdots, R_g^\phi(f)_{q}\right\},\\
\end{equation*}
i.e.,
\begin{equation*}
\psi(g,s,p)\leq \psi(g,s,q).\\
\end{equation*}
$\hfill\blacksquare$

\section{Proof of theorem 2}

\begin{equation*}
\begin{split}
\psi(g,s,m)&=\psi(g,s,m+j),\\
min(\psi(g,s,m))&\geq max(R_g^{\phi}(f)_{b_{m+1}}) \\
\end{split}
\end{equation*}

\noindent\textbf{Proof:}

\noindent From Definition 1, we have
\begin{equation*}
\lim\limits_{m\rightarrow\infty}\psi(g,s,m)=\vee_{s\leq i\leq m}\left\{R_{g}^{\phi}(f)_{b_{i}}\right\}
\end{equation*}
\begin{equation*}
\begin{split}
&=\vee_{s\leq i\leq m}\left\{R_{g}^{\phi}(f)_{b_{i}}\right\}\vee\left\{R_{g}^{\phi}(f)_{b_{m+1}},R_{g}^{\phi}(f)_{b_{m+2}},\cdots,R_g^{\phi}(f)_{b_{\infty}}\right\}\\
&=\psi(g,s,m)\vee\left\{R_{g}^{\phi}(f)_{b_{m+1}},R_{g}^{\phi}(f)_{b_{m+2}},\cdots,R_{g}^{\phi}(f)_{b_{\infty}}\right\}
\end{split}
\end{equation*}
Since $b_{m}\subseteq b_{m+1}\subseteq\cdots\subseteq b_{m+j} $ and $R_g^{\phi}(f)_{b_\infty}=min(g)$ from Appendix A, we get
\begin{equation*}
max(R_{g}^{\phi}(f)_{b_{m+1}})\geq max(R_{g}^{\phi}(f)_{b_{m+2}})\geq\cdots\geq R_{g}^{\phi}(f)_{b_{\infty}}.
\end{equation*}
We have known that $min(\psi(g,s,m))\geq max(R_{g}^{\phi}(f)_{b_{m+1}})$, thus
\begin{equation*}
\psi(g,s,m)\geq\vee\left\{R_{g}^{\phi}(f)_{b_{m+1}},R_{g}^{\phi}(f)_{b_{m+2}},\cdots,R_{g}^{\phi}(f)_{b_{\infty}}\right\},
\end{equation*}
i.e.,
\begin{equation*}
\psi(g,s,m)=\psi(g,s,m+j), \text{where }  m, j\in N^+.
\end{equation*}
$\hfill\blacksquare$

\vfill
\enlargethispage{-5in}
\end{document}